\documentclass[sigconf,nonacm=true,urlbreakonhyphens=true,balance=true]{acmart}
\AtBeginDocument{%
  \providecommand\BibTeX{{%
    \normalfont B\kern-0.5em{\scshape i\kern-0.25em b}\kern-0.8em\TeX}}}

\copyrightyear{2021}
\acmYear{2021}
\setcopyright{acmcopyright}\acmConference[MM '21]{Proceedings of the 29th ACM International Conference on Multimedia}{October 20--24, 2021}{Virtual Event, China}
\acmBooktitle{Proceedings of the 29th ACM International Conference on Multimedia (MM '21), October 20--24, 2021, Virtual Event, China}
\acmPrice{15.00}
\acmDOI{10.1145/3474085.3475535}
\acmISBN{978-1-4503-8651-7/21/10}

\usepackage{multirow}
\usepackage{soul}
\usepackage{colortbl}
\usepackage{enumitem}
\newcommand{\sys}{CoReD}

\begin{document}
\fancyhead{}

%%
%% The "title" command has an optional parameter,
%% allowing the author to define a "short title" to be used in page headers.
% \title{Generalizing Deepfake Detection using Knowledge Distillation}
\title{\sys: Generalizing Fake Media Detection with Continual Representation using Distillation}

\author{Minha Kim, Shahroz Tariq}
% \authornote{Both authors contributed equally to this research.}
% \author{Shahroz Tariq}
% \authornotemark[1]
\affiliation{%
  \institution{College of Computing and Informatics\\ Sungkyunkwan University, South Korea}
  \city{}
  \state{}
  \country{}
}\email{{kimminha,shahroz}@g.skku.edu}

\author{Simon S. Woo}
\authornote{corresponding author}
\affiliation{%
  \institution{Department of Applied Data Science\\
Sungkyunkwan University, South Korea}
  \city{}
    \state{}
  \country{}
}\email{swoo@g.skku.edu}

%%
%% By default, the full list of authors will be used in the page
%% headers. Often, this list is too long, and will overlap
%% other information printed in the page headers. This command allows
%% the author to define a more concise list
%% of authors' names for this purpose.
% \renewcommand{\shortauthors}{Trovato and Tobin, et al.}

%%
%% The abstract is a short summary of the work to be presented in the
%% article.
\begin{abstract}
% As the GAN-based face generation methods have been introduced, a large number of manipulated videos and images with known as DeepFakes have been uploaded constantly on the internet and social media. Furthermore, as the deepfake videos and images have become more sophisticated, many deepfake detection methods have been proposed. However, their performance suffers whenever the emerging
% of deepfakes generation techniques. To detect the new types of deepfake methods while maintaining the prior performance, the deep learning model should be trained with source data.
% However, it is not easy to re-train with all previous data and new data whenever the deepfakes generation methods emerge.
% In this work, we propose the Continual Learning with Representation Learning and Knowledge Distillation paradigms to detect the deepfakes and prevent forgetting knowledge of the existing model without the source data. ??? utilizes

% without forgetting its prior information trained, the deep learning model should be trained with new data and 

Over the last few decades, artificial intelligence research has made tremendous strides, but it still heavily relies on fixed datasets in stationary environments. Continual learning is a growing field of research that examines how AI systems can learn sequentially from a continuous stream of linked data in the same way that biological systems do. Simultaneously, fake media such as deepfakes and synthetic face images have emerged as significant to current multimedia technologies. Recently, numerous method has been proposed which can detect deepfakes with high accuracy. However, they suffer significantly due to their reliance on fixed datasets in limited evaluation settings.  Therefore, in this work, we apply continuous learning to neural networks' learning dynamics, emphasizing its potential to increase data efficiency significantly. We propose Continual Representation using Distillation (\sys) method that employs the concept of Continual Learning (CL), Representation Learning (RL), and Knowledge Distillation (KD). We design CoReD to perform sequential domain adaptation tasks on new deepfake and GAN-generated synthetic face datasets, while effectively minimizing the catastrophic forgetting in a teacher-student model setting. Our extensive experimental results demonstrate that our method is efficient at domain adaptation to detect low-quality deepfakes videos and GAN-generated images from several datasets, outperforming the-state-of-art baseline methods.

% As GAN-based video and image manipulation technologies become more sophisticated and easily accessible, there is an urgent need for effective deepfake detection technologies. Moreover, various deepfake generation techniques have emerged over the past few years. While many deepfake detection methods have been proposed, their performance suffers from new types of deepfake methods on which they are not sufficiently trained. To detect new types of deepfakes, the model should learn from additional data without losing its prior knowledge about deepfakes (catastrophic forgetting), especially when new deepfakes are significantly different. In this work, we employ Continual Learning (CoL), Representation Learning (ReL), and Knowledge Distillation (KD) paradigms to introduce Continual Representation using Distillation (CoReD) method. We use CoReD to perform sequential domain adaptation tasks on new deepfake datasets while minimizing the catastrophic forgetting. Our student model can quickly adapt to new types of deepfake by distilling knowledge from a pretrained teacher model and applying transfer learning without using source domain data during domain adaptation. Through experiments on FaceForensics++ datasets, we demonstrate that CoReD outperforms all baselines on the domain adaptation task with up to 86.97\% accuracy on low-quality deepfakes.
\end{abstract}


%% A "teaser" image appears between the author and affiliation
%% information and the body of the document, and typically spans the
%% page.

\begin{teaserfigure}
\centering
  \includegraphics[width=1\textwidth]{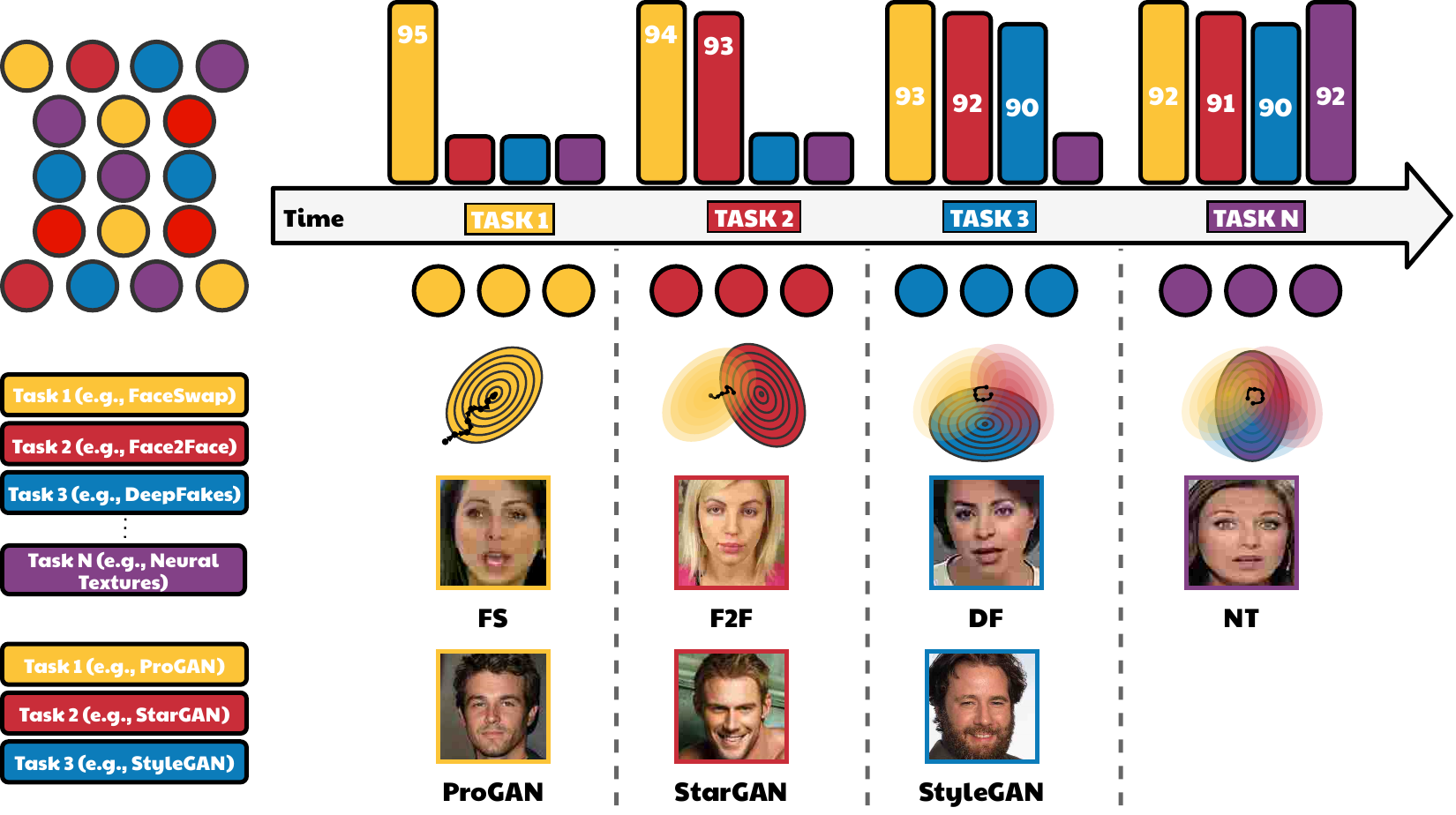}
  \caption{Illustration of our \sys~ using continual representation learning to effectively detect different fake media datasets.}
  \label{fig:teaser}
\end{teaserfigure}
% \begin{teaserfigure}
%   \includegraphics[width=\textwidth]{figures/Pipeline.pdf}
%   \caption{\hl{fix later} \textbf{The architecture of our Feature Representation Transfer Adaptation Learning (FReTAL)}. The teacher model is trained with the Xception.  Before transfer learning, we set the teacher model as untrainable. Then, we initialize the student model with the weights of the teacher model. Target domain data is provided to both teacher and student models to calculate the features for feature storage. We set the teacher as untrainable so that these features are fixed throughout the whole process. Whereas for the student, they will change in each iteration as training progress. We calculate KD loss between teacher and student models and a separate cross-entropy loss function just for the student model. Here, D1-D5 represents the square distance between feature storage of teacher and student.
%     \textit{Note: for the first iteration of transfer learning, the teacher and student model will be the same.}}
%   \label{fig:pipeline}
% \end{teaserfigure}

%%
%% This command processes the author and affiliation and title
%% information and builds the first part of the formatted document.
\maketitle

\section{Introduction}
\label{sec:Intro}

The ability to mimic human learning is desirable and it is a popular criterion 
for artificial intelligence (AI) performance. We measure people's ability to identify pictures, play games, and drive a car and then create machine learning models that can match or exceed the training data provided. While these machine learning models emphasize the final result over the learning process, often they ignores a key feature of human learning such as its robustness and adaptability towards evolving tasks and sequential learning. On the other hand, this robustness stands in sharp contrast to the most efficient state-of-the-art deep learning models, which generally tend to excel, when carefully supplied with a large amount of shuffled, balanced, and homogeneous data. These models not only underperform when faced with slightly different data distributions, but also they fail or experience sharp performance degradation on previously learned tasks, a phenomenon known as catastrophic forgetting~\cite{catastrophic1}. One area where we can clearly observe this phenomenon is fake multimedia detection, especially for deepfake video and GAN-generated image detection, as numerous methods have been researched to generate those~\cite{PGGAN,StarGAN,StyleGAN,VideoGAN_DVDGAN,VideoGAN_TGAN-f,VideoGAN_TGANv2,Deepfakes,FaceApp,FaceSwap,NeuralTextures}. 

Recently, these types of synthetic multimedia from the advanced AI systems are becoming more widespread in social media and online forums~\cite{news1,news2,news3,news4} for creating fake news and information. The recent progress made in deep learning technologies have greatly assisted in generating synthetic images and videos that look strikingly similar to real-world images and videos. Moreover, a large number of fake image generation tools, such as FaceApp~\cite{FaceApp}, FakeApp~\cite{FakeApp} and ZAO~\cite{ZAO}, are also available, which aggravates the situation. It is no secret that deepfakes can severely harm multimedia technologies. However, interested readers can refer to this report on the impact of deepfakes on commercial face recognition APIs~\cite{ShahrozAPI}.
The fake multimedia is generally present in two forms on the Internet: (1) deepfake videos~\cite{FaceForensics++,CelebDF,WildDeepfake} and (2) GAN-generated synthetic images~\cite{StyleGAN,PGGAN,StarGAN} (\textit{From now on, we will refer to them as deepfake videos and GAN images for brevity}). 
On the other hand, there are many fake media detection methods proposed in recent years, achieving state-of-the-art performance~\cite{Shahroz1,MesoNet,Shahroz2,Shahroz3,Hyeonseong3,Hyeonseong2,SAMGAN,FaceForensics++,HASAMCVPRW}. However, as shown by Tariq et al.~\cite{CLRNet}, they suffer from the same robustness and generalization issues when evaluated with a different data distribution than the training set.

More specifically, catastrophic forgetting (knowledge forgetting) problem~\cite{catastrophic1,catastrophic2,Catastrophic3,Catastrophic4} from domain shifting predominantly occurs in such cases. Specifically, catastrophic forgetting is the phenomenon in which deep learning models tend to entirely, partially, or abruptly forget previously learned knowledge upon learning a new task. For example, a classifier's performance trained with one deepfake dataset generally suffers when tested with a different unseen deepfake dataset. Therefore, catastrophic forgetting should be minimized, while transferring information or learning new tasks. 

On the other hand, in order to maintain the knowledge during transferring knowledge, Marra et al.~\cite{incremental_gan} and Rebuffi et al.~\cite{rebuffi2017icarl} proposed approaches to mitigate the data distribution shifting of models by reusing the source data.
By using a feature classifier based on a nearest-neighbor scheme, distribution shift problem can be alleviated. However, due to store the source data to transferring knowledge, they suffer from the limitation of memory resources.

Furthermore, to overcome catastrophic forgetting, Tariq et al.~\cite{CLRNet} use a few data samples from the source domain during transfer learning. However, in practice, for most pre-trained models, either the source domain data is not available or retaining source domain data may raise privacy concerns. Therefore, to encourage maximum applicability in real-world scenarios, we only use the target domain's data and apply knowledge distillation to learn from the pre-trained model (teacher).

Lately, knowledge distillation (KD)~\cite{hinton2015distilling} has shown great success for continual and lifelong learning scenarios~\cite{COL_KD1,COL_KD2}. Motivated by this, we propose, CoReD, a continual learning (CL) based approach using representation learning (RL) and knowledge distillation. In particular, we significantly improve catastrophic forgetting by performing continual learning combined with representation learning and knowledge distillation. 
We use our CoReD method to continuously learn new tasks. Specifically, we use the student loss, distillation loss and representation loss to minimize catastrophic forgetting in the learned task and effectively learn new tasks in a sequential manner to detect a varity of deepfakes at once. Figure~\ref{fig:teaser} provides an overview of our problem settings. A step-by-step illustration of the whole continual learning process is provided in Figure~\ref{fig:pipeline}. We evaluate our method by running experiments on standard GAN image datasets such as ProGAN, StarGAN, and StyleGAN, as well as deepfake video datasets from FaceForensics++ (i.e., DeepFakes, FaceSwap, Face2Face, and NeuralTextures). We demonstrate that CoReD can outperform baseline on low, high, and mixed compression qualities. We find that the teacher-student network and representation loss in our proposed method makes continual learning more effective for deepfake detection. The novel contributions of our work are summarized as follows:

\begin{itemize}[leftmargin=10pt]
    \item \textbf{Continual Representation using Distillation (CoReD). } We propose a novel continual learning framework to detect deepfakes, where we leverage representation learning (RL) and knowledge distillation (KD) with continual learning (CL). %based approach using representation learning (RL) and knowledge distillation
    
    \item \textbf{Extensive Empirical analysis. } We conducted an extensive evaluation with 7 different datasets, including popular GAN-based and deepfake benchmark datasets, using 7 different detection methods and 3 different incremental learning schemes.
    
    \item \textbf{Generalization Performance. } Our approach outperforms other deepfake classification baselines, learning much better on new tasks via continual learning and effectively preventing catastrophic forgetting.
\end{itemize}
Our code is available here\footnote{\url{https://github.com/alsgkals2/CoReD}}.

\section{Related Works}
\label{sec:Related}

In this section, we present the related works on fake multimedia but we mainly focus on visual approaches such as videos and images. Fake speech or text generation is not the main focus of current work. We also overview the relevant research in continual and representation learning as well as knowledge distillation.

\subsection{Fake Image and Video Multimedia}
\subsubsection{\textbf{GAN Image Generation}} 
Researchers in the computer vision community have recently been striving to achieve photorealism for computer-generated face images. Lu et al.~\cite{SurveyFaceSynthesis} provide an overview of these methods. In particular GAN-based methods are also used to alter skin color~\cite{FF46}, generate new viewpoints~\cite{FF34}, and change the age of the face~\cite{FF7}. Also, Fader~\cite{FF43} Networks and Deep Feature Interpolation~\cite{FF62} show promising results on manipulating facial features such as smile, age, and mustache. Moreover, Generative Adversarial Networks (GAN) based methods such as StyleGAN~\cite{StyleGAN}, ProGAN~\cite{PGGAN}, and StarGAN~\cite{StarGAN} are the most famous for generating synthetic faces from scratch. 

However, most of the GAN-based methods have shown to be suffered from low image resolution fidelity; however, Karras et al.~\cite{PGGAN} have improved the image quality using the Progressive growth of GANs (ProGAN). In addition, StyleGAN~\cite{StyleGAN} is an extension of ProGAN architecture to control the style properties of the images. Furthermore, StarGAN~\cite{StarGAN} was proposed to use a single model to perform image-to-image translations for multiple facial
properties. Recently, Variational AutoEncoder (VAE) based methods such as VQ-VAE-2~\cite{VQVAE2} are becoming popular for face synthesis tasks. Like GANs, however, VAE based methods also suffered from low Image fidelity; however, Razavi et al.~\cite{VQVAE2} have proposed a method to generate high-fidelity images using a Vector Quantized Variational AutoEncoder. In this work, we use the more popular ProGAN, StyleGAN, and StarGAN datasets for evaluation and benchmarking different approaches.

\subsubsection{\textbf{Deepfake Video Generation}}
Deepfake video generation techniques have recently gained widespread popularity, and unfortunately, they have been frequently misused to create pornographic videos, fabricated images, and fake news. In particular, in the online communities, recently, there has been a flood of deepfakes or AI-generated synthetic videos~\cite{news1,news2,news3,news4,news5}. In fact, various methods to generate deepfake videos have been proposed. The most viral ones are fabricated using deep learning-based tools such as FakeApp~\cite{FakeApp}, FaceApp~\cite{FaceApp}, and ZAO~\cite{ZAO}. Also, R\"ossler et al.~\cite{FaceForensics++} released the FaceForensics++ (FF++) dataset to advance the research in deepfake detection. The FF++ dataset includes Face2Face (F2F)~\cite{Face2Face} and FaceSwap (FS)~\cite{FaceSwap}, DeepFakes (DF)~\cite{Deepfakes}, and NeuralTextures (NT)~\cite{NeuralTextures}. In this work, we will use the aforementioned four datasets in FaceForensics++ to evaluate our continual learning framework and compare ours with other approaches on learning new deepfake detection tasks.

\subsubsection{\textbf{Fake Image and Video Multimedia Detection}}
Even for human eyes, determining whether images or videos are produced from GAN-based approaches is becoming a more challenging task because of the significant advancements in such generation methods.
In recent years, a plethora of classification methods has been designed to detect deepfake videos~\cite{FaceXRay,CLRNet,MesoNet,FaceForensics++,FaceForensics++,HASAMCVPRW,DeepfakeDetection1,DeepfakeDetection2,DeepfakeDetection3,DeepfakeDetection4,ShahrozAPI,DeepfakeDetection5,DeepfakeDetection6,DeepfakeDetection11,DeepfakeDetection12,Minha_FReTAL,DeepfakeDetection10,SAM_TAR} and GAN images~\cite{Shahroz1,Shahroz2,SAMGAN,transferlearning_tgd}. In particular, Mirsky and Lee~\cite{WenkeSurvey} provide a detailed survey on categorizing different types of detection methods. However, very few of them consider the problem generalizable detectors. Tariq et al.~\cite{CLRNet} demonstrate that many state-of-the-art detectors would fail, when 1) new or unseen deepfakes are tested or 2) training and testing deepfake data are different. They all clearly exhibit catastrophic forgetting. 
Moreover, continual learning for detecting deepfake videos and GAN images has not been thoroughly explored in the past, which is the main focus of our paper. 

Furthermore, we focus on detecting low and mixed-compression quality images that can more frequently occur in real-world social media platforms, whereas much research has focused on detecting high resolution deepfakes~\cite{CLRNet,Shahroz2,Shahroz1}.
In this work, we formulate detecting different types of deepfakes and GAN-based images as learning a new task via continual learning to improve the catastrophic forgetting, enhancing the fundamental generalizability of model and robustness.

\subsection{Continual Representation Learning}
Continual Learning (CL)~\cite{CoL_Survey,CoL_Survey2}, also known as life-long learning, is based on the concept of learning continuously and adaptively. In particular, Continual Learning is a kind of general online learning framework that learn from a infinite stream of data.
Especially, several CL methods~\cite{CL_Catastrophic1,CL_Catastrophic2}  have been introduced to solve the problem of catastrophic forgetting and adjust to dynamically changing tasks. Moreover, CL systems have shown the capability to adapt to and perform well on the entire datasets without revisiting all previous data at each training stage.
Such advantages of CL can tackle key limitations that are prevalent in deep learning and machine learning for generalization and a new task learning. For example, over time, the trained model generally suffers from covariates and knowledge shifts due to vastly and gradually increased size of new datasets, which is also known as \textit{catastrophic forgetting}. Kirkpatrick et al.~\cite{catastrophic2} proposed a constraint-based approach called elastic weight consolidation (EWC) to alleviate catastrophic forgetting in neural networks by selectively restraining the plasticity of weights depending on the importance of weights to previous tasks.
However, their approach shows a lack of scalability because the network size scales quadratically in respect to the number of tasks. In this work, we propose a method that prevents forgetting of knowledge by referring the features of the target data without constraints and the source data.

On the other hand, representation learning (RL) is an approach of learning underlying representations of input data, through transforming or extracting features from data, in order to render machine learning tasks easier to perform. 
Recent research by Long et al.~\cite{re1} explored transferable representation learning with deep adaptation networks to improve the feature transferability in domain adaptation tasks. Their approach embeds the deep features of all task-specific layers into kernel Hilbert spaces (RKHSs), matching optimum domain distributions forming minimax game. However, their approach has not been designed for continual learning setup, which we focus on. In addition, representation learning has been applied to facial expression recognition (FER) task, where Kim et al.~\cite{re2} proposed the new spatio-temporal feature representation learning, which is robust to expression intensity variations. However, their approach only considered two different datasets, which may not be sufficient to assess generalization performance.

\subsubsection{\textbf{Continual Learning using Knowledge Distillation}} 
A pioneering Knowledge distillation (KD) was first proposed by Hinton et al.~\cite{hinton2015distilling} in order to compress and transfer knowledge of a large (teacher) model to a small (student) model. The essence of KD training process is for a student model to effectively mimic the capability of a teacher model. From the continual learning task, Li and Hoiem~\cite{li2017learning} proposed Learning without Forgetting framework to improve catastrophic forgetting by leveraging the knowledge distillation (KD) during transfer learning. 
In addition, to address catastrophic forgetting in class-incremental learning, \textit{rehearsal}~\cite{chen2019catastrophic} principle as well as KD loss were proposed by Rebuffi et al.~\cite{rebuffi2017icarl}. Their work stores the exemplars in the source task to prevent the complete forgetting of a source task. But, for complex inputs, this approach typically requires a very large memory storage to store the features of the source domain. 
To mitigate such a large space requirement, our proposed~\sys~based on continual and representation learning leveraging KD is designed such that it does not require storing or using source exemplars during a new task learning. 

Recently, Hou et al.~\cite{COL_KD1} proposed a novel approach for the multi-task lifelong learning by seeking a better balance between preservation and adaptation via ``Distillation and Retrospection.'' Their proposed CNN-based approach not only helps the
learning on the new task, but also preserves the performance on
old tasks. In particular, Retrospection~\cite{COL_KD1} is proposed to cache a small subset of data for old tasks, which proves to be greatly helpful for the performance preservation, especially in long sequences of tasks drawn from different distributions. We take the similar approach as theirs. However, our approach focuses on continual representation, which was not extensively explored in this work. 
\section{our Approach}
\label{sec:Approach}

\begin{figure*}[t!]
    \centering
    \includegraphics[trim={30pt 15pt 15pt 15pt},clip,width=1\linewidth]{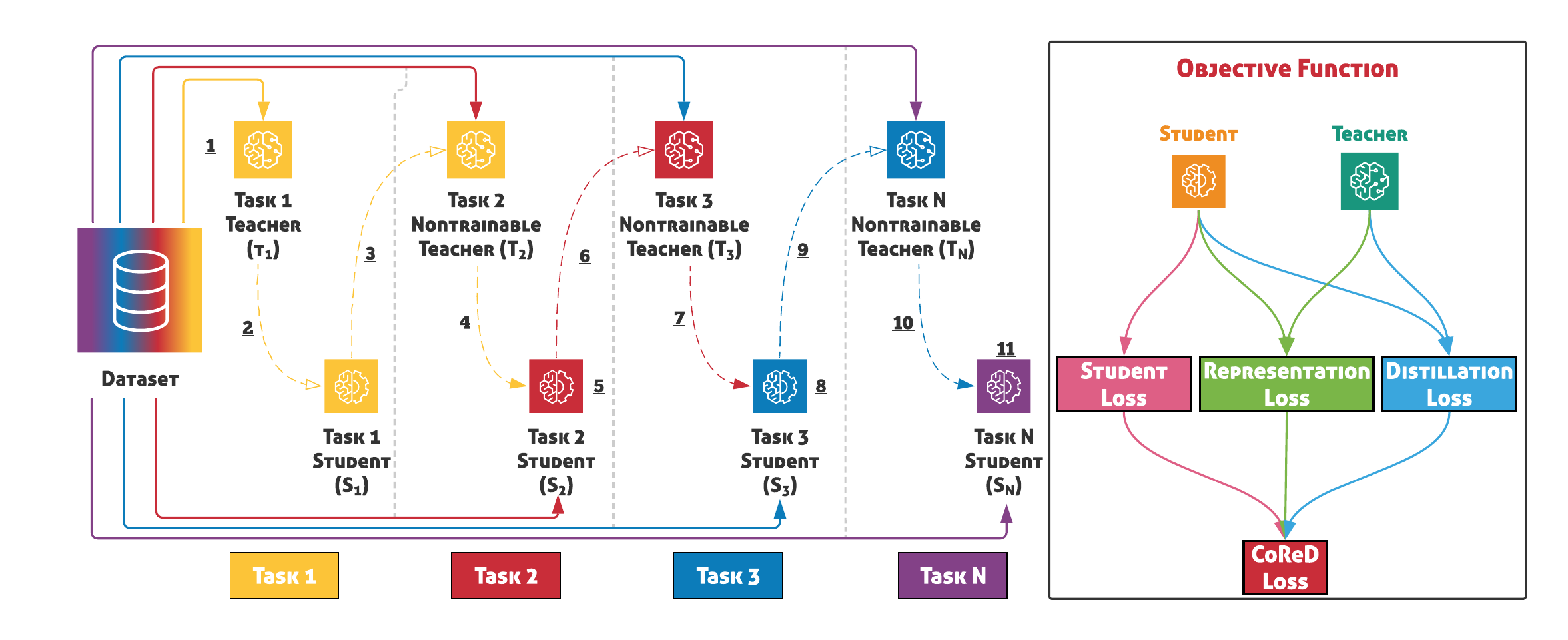}
    \caption{The architecture of our Continual Representation using Distillation (CoReD) method is presented on the left side. The detailed workflow of CoReD is illustrated in steps 1 to 11. The objective function of CoReD is illustrated on the right side using cross entropy, representation, and distillation loss.}
    \label{fig:pipeline}
\end{figure*}

As many new deepfake videos (or GAN images) generation methods are introduced, detecting all of them is becoming more challenging and time-consuming. Therefore, a continual learning-based solution can be beneficial, especially when the data distribution contains an overlap between different generation methods, as in the case of deepfake videos (or GAN images). Consequently, we propose CoReD to effectively detect fake media from various generation methods by using continual learning in a teacher-student model setting. Figure~\ref{fig:pipeline} summarize the overall workflow of CoReD. 

\subsection{CoReD for Fake Multimedia Detection}

Given a deepfake video ($\mathcal{X}_d$) or GAN image ($\mathcal{X}_g$) from any generation method, our goal is to classify it as real or fake. The overall pipeline of our proposed approach is presented in Figure~\ref{fig:pipeline} from Step 1 to Step 11. However, we only describe the first 6 steps from Step 1 to 6, shown in yellow, since the follow-up processes are just repetition of these steps. Therefore, we describe the first 6 steps as follows: 
\begin{enumerate}[leftmargin=13pt]
    \item We first fully train a teacher model $\mathcal{T}_1$ using Task $1$ dataset.
    \item We copy the weights from the teacher trained on task $1$ into the student model ($\mathcal{S}_1$).
    \item Now, we change the student into the teacher for Task $2$ ($\mathcal{T}_{2}$) and set $\mathcal{T}_{2}$ as non-trainable.
    \item Next, we create a new student model ($\mathcal{S}_2$) by copying the weight from Task $2$ teacher $\mathcal{T}_{2}$, setting $\mathcal{S}_2$ as trainable.
    \item Now, we supply $\mathcal{T}_{2}$ and $\mathcal{S}_2$ with the data from Task $2$. The student learns from the data in the following three ways: (a) directly from the data using a cross-entropy loss (student loss $\mathcal{L}_{\mathcal{S}}$), (b) using a representation loss calculated between $\mathcal{T}_{2}$ and $\mathcal{S}_2$ by comparing the feature representation memory (representation loss $\mathcal{L}_{\mathcal{R}}$), and (c) using knowledge distillation loss calculated using the $\mathcal{T}_{2}$ and $\mathcal{S}_2$ (distillation loss $\mathcal{L}_{\mathcal{D}}$), where we describe the details of each loss in next section.
    
    Note: Since $\mathcal{T}_{2}$ is set as non-trainable, its feature representation memory and knowledge remain the same. However, for $\mathcal{S}_2$, they will gradually change over the training epochs (see Section~\ref{sec:objective} for details on the objective function).
    \item Once the student ($\mathcal{S}_2$) is fully trained, we go back to Step (3) and use it as the teacher (i.e., $\mathcal{T}_{3}$) for the next task, and we repeat this process until we finish Task $N$ (i.e., Step 11 in Figure~\ref{fig:pipeline}).
\end{enumerate}
We provide a detailed step-by-step overview of this process in Figure~\ref{fig:pipeline}.

\subsection{Objective Function}
\label{sec:objective}
We calculate three loss functions for CoReD, as shown in Figure~\ref{fig:pipeline}. The details are presented as follows:

\textbf{Student Loss. } As we mentioned earlier, when we train the student model ($\mathcal{S}$), as shown in steps 5, 8, and 11 in Figure~\ref{fig:pipeline}, we use the cross-entropy loss to learn directly from the Task's dataset as follows:
\begin{equation}
\small
    \begin{split}
         \sigma(s)_i&=\frac{e^{s_i}}{\sum_j^C e^{s_j}},\\
        \mathcal{L}_{\mathcal{S}}&= -{\sum_{i=1}^{C=2} t_i\log(\sigma(\mathcal{S}(x_i,y_i))_i)} \\&=-t_1\log(\sigma(\mathcal{S}(x_i,y_i))_1)-(1-t_1)\\
        &\quad\quad \log(1-\sigma(\mathcal{S}(x_i,y_i))_1),
    \end{split}
    \label{eq:Studentloss}
\end{equation}
where $\sigma$ is the softmax function, $C_1$ and $C_2$ are the real and fake classes: $t_1$[0,1] and $\sigma(\cdot)_1$ are the ground truth and the score for $C_1$, and $t_2=1-t_1$ and $\sigma(\cdot)_2= 1-\sigma(\cdot)_1$ are the ground truth and the score for $C_2$. Moreover, $y_i$ is the output label (i.e., hard label $y$), and $\hat y_i$ is the output of $\mathcal{S}$ (i.e., hard prediction). 

\textbf{Distillation Loss. }
During the student model training at Step 5, 8, and 11 in Figure~\ref{fig:pipeline}, we also calculate the distillation loss using the student from the teacher model as follows:
\begin{equation}
 \small
    \begin{split}
        \sigma_d(s,T)_i &= \frac{e^{(\frac{s_i}{T})}}{\sum_{j}^{C} e^{(\frac{s_j}{T}))}},\\
        \mathcal{L}_{\mathcal{D}}&=\sum_{x_i\in \mathcal{X}} \mathcal{L}(\mathcal{T}(x_i),\mathcal{S}(x_i))\\
        &=\sum_{x_i\in \mathcal{X}} \sigma_d(\mathcal{T}(x_i,y_i);T=\tau) \log \sigma_d(\mathcal{S}(x_i,\hat y_i);T=\tau)),\\
    \end{split}
    \label{eq:Distillationloss}
\end{equation}
where $\sigma_d$ is the softmax function with temperature $T$ initialized as $\tau$ during distillation. And, $y_i$ is the output label of $\mathcal{T}$ (i.e., soft label $y$), and $\hat y_i$ is the output of $\mathcal{S}$ (i.e., soft prediction). By softening the probability distribution over the classes, the temperature aids $\mathcal{S}$ in imitating $\mathcal{T}$. By increasing $T$, the probability distribution of the softmax function softens, revealing which classes $\mathcal{T}$  is more similar to the predicted class.

\textbf{Representation Loss. }We believe that similar or common underlying characteristics must exist among different types of fake multimedia (deepfake video or GAN images) generated from various generation methods. Therefore, the teacher ($\mathcal{T}$) trained on Task $i$ can help the student ($\mathcal{S}$) learn Task $i+1$ using fewer samples. Therefore, during the training of the student model, we store the feature representations of $\mathcal{T}$ and $\mathcal{S}$ for the training data in a representation memory ($\mathcal{R}_{mem.}$). Rather than storing all Task $i+1$ data features, we only store distinct features selectively to minimize the memory footprints, unlike storing a large number of samples from prior research~\cite{rebuffi2017icarl}. 

To accomplish this, we apply softmax to the output of $\mathcal{T}$ and $\mathcal{S}$, which we use to make representation memory $\mathcal{R}_{mem.}^\mathcal{T}$ and $\mathcal{R}_{mem.}^\mathcal{S}$. We partition this representation memory into small blocks $b$ each of size $v$ starting from $m$ as follows: $\mathcal{R}_{mem.} =\{(m, m+v),(m+v,m+2v),\dots,(m+ (b-1)v,m +bv)\}$. Dividing the memory helps reduce context shifting during the learning process. Since the distribution of real and fake data are different, we perform this operation separately on both real and fake data. 
Finally, we calculate the difference between $\mathcal{R}_{mem.}^\mathcal{T}$ and $\mathcal{R}_{mem.}^\mathcal{S}$, as follows:
\begin{equation}
    \mathcal{L}_{\mathcal{R}}=\sum_{1}^{b} \left \| \mathcal{R}_{mem.(b)}^\mathcal{S}-\mathcal{R}_{mem.(b))}^\mathcal{T} \right \|_2^2,
    \label{eq:Representationloss}
\end{equation}
where for binary classification, we divide the feature storage into $b=5$, each of size $v=0.1$ start from $m=0.5$, e.g., $\mathcal{R}_{mem.(1)}^\mathcal{S}$ is the first block of student's representation memory.

\textbf{Final CoReD Loss .}We take the sum of all three losses from Equation~\eqref{eq:Studentloss}, \eqref{eq:Distillationloss}, and \eqref{eq:Representationloss} to form our final CoReD loss function, as follows:
\begin{equation}
    \mathcal{L}_{CoReD}= \alpha \mathcal{L}_{\mathcal{S}}+\beta \mathcal{L}_{\mathcal{D}}+\gamma \mathcal{L}_{\mathcal{R}},
    \label{eq:CoReDloss}
\end{equation}
where $\alpha$, $\beta$, and $\gamma$ are coefficients to control the three loss terms.
\section{Experiment}
\label{sec:Experiment}

\begin{table}[t!]
\centering
\caption{Dataset Description. The details of the total number of images used for training and testing.}
\vspace{-10pt}
% \vspace{-10pt}
\label{tab:dataset_details}
\resizebox{1\linewidth}{!}{%
\begin{tabular}{lcccc} 
\toprule
\multicolumn{1}{l}{\textbf{Datasets} } & \begin{tabular}[c]{@{}c@{}}\textbf{Total}\\\textbf{Images} \end{tabular} & \begin{tabular}[c]{@{}c@{}}\textbf{Task 1}\\\textbf{Images} \end{tabular} & \begin{tabular}[c]{@{}c@{}}\textbf{Continual}\\\textbf{Learning} \end{tabular} & \begin{tabular}[c]{@{}c@{}}\textbf{Testing}\\\textbf{Images} \end{tabular} \\ 
\hline
Pristine (Real) & 80,000 & 60,000 & 800 & 20,000 \\ 
\hline
DeepFake (DF) & 80,000 & 60,000 & 800 & 20,000 \\ 
\hline
FaceSwap (FS) & 80,000 & 60,000 & 800 & 20,000 \\ 
\hline
Face2Face (F2F) & 80,000 & 60,000 & 800 & 20,000 \\ 
\hline
NeuralTextures (NT) & 80,000 & 60,000 & 800 & 20,000\\ 
\hline
ProGAN & 60,000 & 18,000 & 2,000 & 20,000 \\ 
\hline
StyleGAN & 65,838 & 13,800 & 2,000 & 60,000 \\ 
\hline
StarGAN & 274,478 & 30,520 & 2,000 & 100,000 \\ 
\hline
CelebA-HQ & 49,000 & 30,000 & 2,000 & 10,000 \\ 
\bottomrule
\end{tabular}
}
\vspace{-10pt}
\end{table}

\subsection{Dataset}
To evaluate the performance of our approach, we used ProGAN~\cite{PGGAN}, StyleGAN~\cite{StyleGAN}, and StarGAN~\cite{StarGAN} as GAN image datasets as well as DeepFake (DF), Face2Face (FS), FaceSwap (FS), and NeuralTextures (NT) datasets from FaceForensics++~\cite{FaceForensics++} as deepfake video dataset (see supplementary material for CelebDF results). We use pristine videos from FaceForensics++ as real videos and images from the CelebA~\cite{CelebA} dataset as real images. We provide the dataset details in Table~\ref{tab:dataset_details}. We use 60,000 images from 750 videos for Training the teacher on Task 1 for deepfake detection. Then, for every next continual learning task, we 800 images from using just ten videos. The intuition behind using such a small number of samples for continual learning is that obtaining an extensively large dataset for a new type of forgery method is challenging in a real-world environment. Therefore, achieving high detection performance while using a minimum amount of data is crucial, and it is highly desirable. Similarly, we use 18,000 ProGAN images (or 13,800 StyleGAN images or 30,520 StarGAN images) for training the teacher in Task 1, and for every next continual learning task, we use 2,000 images. Also, we use 20,000 images from 125 deepfakes videos. For GAN images, we use 20,000 for ProGAN images, 60,000 for StyleGAN images, and 100,000 for StarGAN, respectively for testing all detection methods. In addition, we perform our evaluation using high, low, and mixed\footnote{We develop the mixed quality dataset by combining low- and high-quality deepfakes datasets. For the training and testing of mixed quality: half of the samples are high-quality ,and half are of low-quality.} compression qualities and report the results in Section~\ref{sec:Results}.

\subsection{Experimental Settings}
We train each method for 100 iterations using early stopping with a patience value of 5. We employ the stochastic gradient descent (SGD) with a momentum of 0.1 and a learning rate of 0.05. We set the value of hyper-parameter values as follows: $\tau=20$, $\alpha=1.0$, $\beta=1.0$, and $\gamma=1.0$. The representation memory is of this form $\mathcal{R}_{mem.} =\{(0.5,0.6),(0.6,0.7),\dots,(0.9,1.0)\}$. Our tests are carried out on Intel Core i7-9700 CPU with 32 GB of RAM and Nvidia RTX 3090 GPU. We use F$_1$-score metric for evaluation. We extract 80 samples from each video. We use the MTCNN~\cite{MTCNN} algorithm for facial landmark detection on real and deepfake videos. Using this information, crop a $128\times128$ square with the face at its center, thereby preserving the aspect ratio of the image. For GAN and CelebA images, the faces are already at the center of the images. Therefore, we only resize them to $128\times128$ resolution. We apply the following normalization settings using PyTorch Transform: [0.5,0.5,0.5]. We use CutMix ~\cite{yun2019cutmix} for our data augmentation.

\textbf{Task Description. }We perform Task 1 to 4 for Deepfake video detection and Task 1 to 3 for GAN image detection. In Task 1, we fully train the backbone model on a single deepfake (or GAN image) dataset. From Task 2 to Task N, we perform continual learning to new deepfake (or GAN image) datasets, as shown in Figure~\ref{fig:pipeline}.

\subsection{Baselines}

We use the following baselines for evaluation on Task 1. Note: we use the best performer among these baselines for Task 2, 3, and 4.
\begin{itemize}[leftmargin=10pt]
    \item \textbf{MesoNet. } Afchar et al.~\cite{MesoNet} proposed MesoNet, which is a CNN-based state-of-the-art deepfake detection method. 
    \item \textbf{ShallowNet. } ShallowNet proposed by Tariq et al.~\cite{Shahroz2} also has shown to be effective for distinguishing real and fake images. 
     \item \textbf{ForensicTransfer. } Cozzolino et al.~\cite{cozzolino2018forensictransfer} proposed Forensic Transfer, a weakly supervised method used for transfer learning to different domains.
    \item \textbf{DBiRNN}: Sabir et al.~\cite{DFD2} a bidirectional RNN with DenseNet and demonstrated high performance on FaceForensics++ dataset. 
    \item \textbf{Xception. } Xception~\cite{chollet2017xception} is considered as the state-of-the-art method for the ImageNet classification task. Also, Rössler et al.~\cite{FaceForensics++} showed that Xception is one of the competivie performers on the FaceForensics++ dataset. 
    \item \textbf{CNN+LSTM. } Güera et al.~\cite{FT_SQ2} proposed a CNN concatenated with LSTM network to detect deepfake.
    \item \textbf{EfficientNet. } Tan et al.~\cite{Efficientnet} proposed EfficientNet and it is considered as the state-of-the-art at several classification tasks. We use EfficientNet-B0 for our evaluation.
\end{itemize}
We use CNN+LSTM, DBiRNN ShallowNet, MesoNet, and Xception for deepfake video detection. Since the code for CNN+LSTM and DBiRNN is not publicly accessible, we implemented them and did our best to fit the experimental settings in the original article. We use MesoNet, ForensicsTransfer, Xception, and EfficientNet-B0 for GAN image detection.

To evaluate our continual learning framework (Task 2 to N), we compare CoReD method with three incremental learning methods. We use the best performer on Task 1 and use it with these methods for the following Task 2, 3, and 4.
\begin{itemize}[leftmargin=10pt]
    \item \textit{\textbf{Transfer Learning (TF)}}: The first method is Transfer learning, where we perform fine-tuning on the model to learning the new Task.
    \item \textit{\textbf{Transferable GAN-images Detection framework (TG)}}: The second method is a KD-based GAN image detection framework using L2-SP and self-training.
    \item \textit{\textbf{Distillation Loss (DL)}}: The third method is a part of our ablation study, where we only use the distillation loss $\mathcal{L}_{\mathcal{D}}$ component from our CoReD loss function ($\mathcal{L}_{CoReD}$) to perform incremental learning.
\end{itemize}
\section{Results}
\label{sec:Results}
This section discusses the deepfake as well as GAN image detection performance for Task 1 to 4 and provides an ablation study.

\begin{table}
\centering
\caption{Task 1 result for GAN image detection. The best performer is highlighted in boldface. Based on the results from this experiment, we select EfficientNet-B0 as the backbone model for Task 2 and 3 for GAN image detection.}
\vspace{-10pt}
\label{tab:task1_GAN}
\resizebox{\linewidth}{!}{%
\begin{tabular}{lccc} 
\toprule
\textbf{Method\ \ \ \ \ \ \ \ \ \ \ \ \ \ \ \ \ \ \ \ \ \ \ \ \ \ \ \ } & \textbf{ProGAN} & \textbf{StyleGAN} & \textbf{StarGAN} \\ 
\hline
MesoNet~\cite{MesoNet} & 97.47 & 87.98 & 99.50 \\
ForensicsTransfer~\cite{cozzolino2018forensictransfer} & 97.15 & 99.12 & 85.34 \\
Xception~\cite{chollet2017xception} & 99.90 & 99.92 & \textbf{99.90} \\
EfficientNet-B0~\cite{Efficientnet} & \textbf{99.91} & \textbf{99.96} & 99.88 \\
\bottomrule
\end{tabular}
}
\vspace{-5pt}
\end{table}

\begin{table}
\centering
\caption{Task 1 result for deepfake video detection. The best performer is highlighted in boldface. Based on the results from this experiment, we select Xception as the backbone model for Task 2, 3, and 4 for deepfake video detection.}
\vspace{-10pt}
\label{tab:task1_deepfake}
\resizebox{\linewidth}{!}{%
\begin{tabular}{lccccc} 
\toprule
\textbf{Method\ \ \ \ \ \ \ \ \ \ \ \ \ \ \ \ \ \ \ \ } & \textbf{DF}  & \textbf{FS}  & \textbf{F2F}  & \textbf{NT}  & \textbf{Avg.} \\ 
\hline
CNN+LSTM~\cite{FT_SQ2} & 78.51 & 77.75 & 71.87 & 90.54 & 77.80 \\
DBiRNN~\cite{DFD2} & 80.54 & 80.56 & 73.12 & 94.38 & 82.21 \\
ShallowNet~\cite{Shahroz1} & 88.97 & 93.33 & 75.26 & 99.45 & 87.08 \\
MesoNet~\cite{MesoNet} & \textbf{99.01} & 99.26 & 99.01 & 99.27 & 99.14\\
Xception~\cite{chollet2017xception} & 99.00 & \textbf{99.29} & \textbf{99.26} & \textbf{99.46} & \textbf{99.25} \\
\bottomrule
\end{tabular}
}
\vspace{-5pt}
\end{table}
\subsection{Performance on Task 1}
In this section, we present the Task 1 performance of the baseline method for GAN image and deepfake video detection. From Table~\ref{tab:task1_GAN}, we can observe that all baseline methods perform exceptionally well at detecting GAN images, where EfficientNet-B0 is the best performer on average. Therefore, we use EfficientNet-B0 for the continual learning process (i.e., Task 2 to 3) of GAN images. On the other hand, Xception is the best performer for deepfake video detection with an average F1-score of 99.25\%, as shown in Table~\ref{tab:task1_deepfake}. Therefore, we use Xception for the following continual learning process (i.e., Task 2 to 4) of deepfake videos.

\begin{table}
\centering
\caption{Zero-shot performance of Task 1's best performer (EfficientNet-B0) on GAN image dataset. Based on these results, it is evident that a continual or incremental learning solution is necessary for the GAN image dataset.}
\vspace{-10pt}
\label{tab:LQ_zeroshot_GAN}
\resizebox{\linewidth}{!}{%
\begin{tabular}{llccc} 
\toprule
\textbf{Method} &  \textbf{Dataset} & \textbf{ProGAN}  & \textbf{StyleGAN}  & \textbf{StarGAN}\\ 
\hline
EfficientNet-B0 & \textbf{ProGAN} & \textbf{99.91} & 49.47 & 56.81    \\
\hline
EfficientNet-B0 & \textbf{StyleGAN} & 49.80 & \textbf{99.96} & 50.04   \\
\hline
EfficientNet-B0 & \textbf{StarGAN} & 66.47 & 52.01 & \textbf{99.88}   \\
\bottomrule
\end{tabular}}
\vspace{-5pt}
\end{table}

\begin{table}
\centering
\caption{Zero-shot performance of Task 1's best performer (Xception) on deepfake video dataset. Similar to the zero-shot results on GAN images, the result on deepfake videos also supports the argument that a continual or incremental learning solution is required.}
\vspace{-10pt}
\label{tab:LQ_zeroshot_deepfake}
\resizebox{\linewidth}{!}{%
\begin{tabular}{llcccc} 
\toprule
\textbf{Method} & \textbf{Dataset} & \textbf{\ \ \ \ DF\ \ \ \ }  & \textbf{\ \ \ \ F2F\ \ \ \ }  & \textbf{\ \ \ \ FS\ \ \ \ }  & \textbf{\ \ \ \ NT\ \ \ \ }  \\ 
\hline
Xception  & \textbf{\ \ \ DF} & \textbf{99.41}  & 56.05 & 49.93 & 66.32 \\ 
\hline
Xception & \textbf{\ \ \ F2F} & 68.55 & \textbf{98.64}  & 50.55 & 54.81 \\ 
\hline
Xception  & \textbf{\ \ \ FS} & 49.89 & 54.15 & \textbf{98.36}  & 50.74 \\ 
\hline
Xception & \textbf{\ \ \ NT} & 50.05 & 57.49 & 50.01 & \textbf{99.88}  \\
\bottomrule
\end{tabular}
}
\vspace{-5pt}
\end{table}

\textbf{Zero-shot Performance. }
We demonstrate that it is essential to perform continual learning for deepfake videos and GAN image detection. We evaluate the best performers of Task 1 for GAN image detection (i.e., EfficientNet-B0) with different GAN image datasets. For example, if ProGAN is used as Task 1 dataset, we evaluate the performance of EfficientNet-B0 using StyleGAN and StarGAN (i.e., Zero-shot performance). As shown in Table~\ref{tab:LQ_zeroshot_GAN}, EfficientNet-B0 fails to detect everything outside its training data distribution, showing a lack of robustness and generalizability. In Table~\ref{tab:LQ_zeroshot_deepfake}, we observe similarly poor zero-shot performance results for Xception at deepfake videos detection. These results demonstrate that off-the-shelf deepfake detectors are only good at detecting one type of forgery or only those which are very similar to their original training data distribution. And a continual learning solution is clearly required so that model can quickly adapt to new data distributions.

\subsection{Continual Learning Results (Task 2-4)}
In this section, we provide the performance results of continual learning (i.e., from Task 2 to N). We use Xception for deepfake video detection and EfficientNet-B0 for GAN image detection as the backbone model throughout this section. Further, we use CoReD and the three continual learning methods with these backbone methods (i.e., TF, TG, and DL) to evaluate continuous learning.

\begin{table}
\centering
\caption{Task 2 results for deepfake video and GAN image detection. The best performer is highlighted in boldface. All methods perform relatively well. However, CoReD shows the best performance.}
\vspace{-10pt}
\label{tab:task2}
\resizebox{\linewidth}{!}{%
\begin{tabular}{lccccc} 
\toprule
\multirow{2}{*}{\textbf{Method}} & \multicolumn{3}{c}{\textbf{Deepfake Videos}} & \multirow{2}{*}{\begin{tabular}[c]{@{}c@{}}\textbf{GAN}\\\textbf{Images}\end{tabular}} & \multirow{2}{*}{\textbf{Average}} \\
 & \textbf{HQ} & \textbf{LQ} & \textbf{Mixed} &  &  \\ 
\hline
TranLearn (TF) & 95.02 & 85.35 & 89.01 & 83.79 & 88.29 \\
TranGAN (TG) & 87.55 & 74.17 & 86.67 & 82.90 & 82.82 \\
DistilLoss (DL) & 74.04 & 85.32 & 91.00 & 56.69 & 76.76 \\
\textbf{CoReD (Ours)} & \textbf{95.15} & \textbf{86.97} & \textbf{92.48} & \textbf{85.36} & \textbf{89.99} \\
\bottomrule
\end{tabular}
}
\vspace{-10pt}
\end{table}

\textbf{Performance on Task 2. }
Now, we examine the performance of Task 2 for deepfake videos and GAN image detection from Table~\ref{tab:task2}. We can observe that CoReD is the best performer (89.99\% on average) for deepfake videos and GAN images. It is also interesting to note that there is no clear winner among the TF, TG, and DL. However, TF demonstrates the most stable performance (88.29\%) among the three continual learning baselines, showing that regular transfer learning works well for the first level of incremental learning. Nevertheless, every method performs relatively well on Task 2. It shows that all four methods (i.e., TF, TG, KD, and CoReD) used for evaluation are well equipped to learn new tasks. Therefore, making the comparison more competitive for the follow-up tasks.

\begin{table}
\centering
\caption{Task 3 results for deepfake video and GAN image detection. TF and TG suffer from catastrophic forgetting, whereas DL and CoReD demonstrate good performance. The best performer is highlighted in boldface. }
\vspace{-10pt}
\label{tab:task3}
\resizebox{\linewidth}{!}{%
\begin{tabular}{lccccc} 
\toprule
\multirow{2}{*}{\textbf{Method}} & \multicolumn{3}{c}{\textbf{Deepfake Videos}} & \multirow{2}{*}{\begin{tabular}[c]{@{}c@{}}\textbf{GAN}\\\textbf{Images}\end{tabular}} & \multirow{2}{*}{\textbf{Average}} \\
 & \textbf{HQ} & \textbf{LQ} & \textbf{Mixed} &  &  \\ 
\hline
TranLearn (TF) & 75.81 & 67.69 & 74.90 & 85.91 & 76.08 \\
TranGAN (TG) & 88.58 & 64.61 & 69.93 & 69.46 & 73.15 \\
DistilLoss (DL) & 93.03 & 86.09 & 75.52 & 83.54 & 84.54 \\
\textbf{CoReD (Ours)} & \textbf{94.09} & \textbf{87.79} & \textbf{77.41} & \textbf{86.44} & \textbf{86.43} \\
\bottomrule
\end{tabular}
}
\vspace{-5pt}
\end{table}

\textbf{Performance on Task 3. }
In this section, we discuss the results of Task 3 from Table~\ref{tab:task3} Now, the performance difference between the baselines and CoReD is distinctively clear. CoReD achieves nearly 10\% better detection performance than TF and TG. And only DL (84.54\%) performs close to CoReD (86.43\%). Also, we observe that the performance on the mixed dataset is relatively lower than the other, exhibiting that they are harder to detect. Overall, we can witness the effect of catastrophic forgetting on the TF and TG as their detection performance drops nearly 15\% from the previous task (see Table~\ref{tab:task2}). Since there is only a small performance drop of less than 5\% on average for CoReD and DL, this confirms that the teacher-student architecture clearly helps resist catastrophic forgetting.

\begin{table}
\centering
\caption{Task 4 results for deepfake video detection. All baseline (i.e., TF, TG, and DL) suffers from catastrophic forgetting, whereas CoReD demonstrates stable and consistent performance. The best performer is highlighted in boldface.}
\vspace{-10pt}
\label{tab:task4}
\resizebox{\linewidth}{!}{%
\begin{tabular}{lcccc} 
\toprule
\multirow{2}{*}{\textbf{Method}\ \ \ \ \ \ \ \ \ \ \ \ \ \ \ \ \ \ \ \ \ \ \ \ \ } & \multicolumn{3}{c}{\textbf{Deepfake Videos}}  \\
 & \textbf{\ \ \ \ HQ\ \ \ \ } & \textbf{\ \ \ \ LQ\ \ \ \ }  &  \textbf{\ \ \ \ Average\ \ \ \ } \\ 
\hline
TranLearn (TF) & 55.97 & 68.53 & 62.25 \\
TranGAN (TG) & 85.46 & 60.45  & 72.96 \\
DistilLoss (DL) & 90.63 & 67.21  & 78.92 \\
\textbf{CoReD (Ours)} & \textbf{91.77} & \textbf{80.55} & \textbf{86.16} \\
\bottomrule
\end{tabular}
}
\vspace{-10pt}
\end{table}

\textbf{Performance on Task 4. }Finally, in this section, we present the results of Task 4 from Table~\ref{tab:task4}. The performance of all baselines (i.e., TF,  TG, and DL) significantly drops on Task 4. The detection performance of TF and TG is below 70\%, exhibiting a significant impact of catastrophic forgetting. In contrast, the detection performance of our CoReD is more than 20\% higher than the best baseline method (i.e., DL) in some scenarios. It demonstrates that in the long run, CoReD is a more stable and consistent performer. The DL method, which was relatively competitive with CoReD and showed resistance to catastrophic forgetting until Task 3. This demonstrates inconsistent detection performance, i.e., DL (HQ: 90.63\% and LQ: 67.21\%) vs. CoReD (HQ: 91.77\% and LQ: 80.55\%). We believe that the collective contribution of the loss functions of CoReD (i.e., $ \mathcal{L}_{\mathcal{S}}$, $\mathcal{L}_{\mathcal{D}}$ and $\mathcal{L}_{\mathcal{R}}$,) assist in providing robustness against catastrophic forgetting.

\subsection{Ablation Study}
In this section, we perform an ablation study on the following aspects of the proposed CoReD: Q1. What is the impact of the block size of the representation memory on detection performance? Q2. What is the impact of removing Representation memory and student loss on CoReD? and Q3. What is the impact of performing continual learning to two tasks at the same time?

\begin{table}
\centering
\caption{Ablation study on block size ($b$) of representation memory ($\mathcal{R}_{mem.}$). We observe a drop in performance by removing the memory block and using a single memory storage ($b=1$).}
\vspace{-10pt}
\label{tab:ablation1}
\resizebox{\linewidth}{!}{%
\begin{tabular}{lccc} 
\toprule
\multirow{2}{*}{\textbf{Method\ \ \ \ \ \ \ \ \ \ \ \ \ \ \ }} & \multicolumn{3}{c}{\textbf{Deepfake Videos}} \\
 & \textbf{\ \ \ \ LQ\ \ \ \ } & \textbf{\ \ \ \ HQ\ \ \ \ } & \multicolumn{1}{l}{\textbf{\textbf{\ \ \ \ Average\ \ \ \ }}} \\ 
\hline
CoReD ($b=1$)\ \ \ \ \ \ \ \ \ \ \ \ \ \ \  & 75.30 & 83.34 & 79.32 \\
\textbf{CoReD ($b=5$)}\ \ \ \ \ \ \ \ \ \ \ \ \ \ \  & \textbf{78.56} & \textbf{89.42} & \textbf{83.99} \\ 
\bottomrule
\end{tabular}
}
% \vspace{-10pt}
\end{table}
\textbf{Q1. Block Size of Representation Memory. }We change the representation memory ($\mathcal{R}_{mem.}$) into one single memory block ($b=1$). We compare it with CoReD using a standard setting (i.e., $b=5$) and present the results in Table~\ref{tab:ablation1}. The detection performance for both low and high-quality deepfakes decreases up to 6\% by setting $b=1$. Using five memory blocks ($b=5$) for CoReD can increase the detection performance by 4\% on average; hence its use is well justified.

\textbf{Q2. Without Representation and Student Loss. } For this ablation study, we consider the scenarios, where we do not use our proposed representation loss ($\mathcal{L}_{\mathcal{R}}$) and student loss ($\mathcal{L}_{\mathcal{S}}$) with CoReD. The results for this setting are summarized in Table~\ref{tab:task2}--\ref{tab:task4} as the DL method. As we discussed earlier, CoReD outperforms DL in most scenarios, demonstrating the use of representation and student loss in our objective function.

\begin{table}
\centering
\caption{Ablation study on continual learning to two tasks simultaneously. We observe a slight increase in performance when learning simultaneously.}
\vspace{-10pt}
\label{tab:ablation2}
\resizebox{\linewidth}{!}{%
\begin{tabular}{l|c|c|c} 
\toprule
\multicolumn{4}{c}{\textbf{Deepfake Videos (HQ)}} \\ 
\hline
\textbf{Method} & \textbf{\ \ \ Task 1$\rightarrow$2\&3\ \ \ } & \textbf{\ \ \ Task 1$\rightarrow$2$\rightarrow$3\ \ \ } & \textbf{\ \ \ Average\ \ \ } \\ 
\hline
DL & 94.17 & 93.03 & 93.60 \\
\textbf{CoReD} & \textbf{94.18} & \textbf{94.09} & \textbf{94.13} \\
\bottomrule
\end{tabular}
 }
\end{table}

\textbf{Q3. Simultaneous Continual Learning to Two Tasks. }Lastly, we compare the learning performance on simultaneous two tasks using continual learning with the typical scenarios of continual task-by-task learning. As shown in Table\ref{tab:ablation2}, simultaneous continual learning results in a slightly higher detection performance than the task-by-task approach, i.e., 94.18\% vs. 94.09\%. However, the difference is not very significant. Moreover, CoReD outperforms DL (the best baseline method) in both scenarios. These results also suggest that both scenarios are practically applicable.
\section{Discussion}
\label{sec:Discussion}

\noindent
\textbf{New Data Augmentation Methods. }In this work, we have not explored many of the recent data augmentation techniques, which have proven to be very successful against deepfake detection, as shown in the deepfake detection challenge~\cite{DFDC}. However, the focus of our current work is more towards the continual learning component. And adding any state-of-the-art data augmentation technique or detection model or preprocessing method in our pipeline will likely increase our overall detection performance. We believe that if CoReD can demonstrate promising detection performance on general classification methods such as Xcpetion and EfficientNet-B0, it will also provide higher detection performance on those classifiers specially built for deepfake detection.

\noindent \textbf{Mixed Quality Detection. } As we know, training dataset with high-quality deepfakes does not perform well against detecting deepfakes of low quality. Accordingly, it is interesting to note that unless we apply the same compression to high-quality deepfakes for converting them into low-quality. The model trained on low-quality deepfakes will also suffer from them. Therefore, to further explore this, we mixed the low and high-quality samples from deepfake datasets named it mixed quality. As shown in Table~\ref{tab:task3}, the performance on mixed quality can sometimes be even lower than low-quality deepfakes. Therefore, more research is needed to improve the detector's performance on mixed quality deepfakes. 

\noindent\textbf{Limitations and Future Work. } In this work, we only evaluate four types of deepfake datasets and three types of GAN image datasets. Recently, several more advanced deepfake datasets (DFDC~\cite{DFDC}, CelebDF~\cite{CelebDF}, CelebFOM~\cite{ShahrozAPI}, and WildDeepfake~\cite{WildDeepfake}) as well as GAN image datasets (CelebBlend~\cite{ShahrozAPI}, GANsformer~\cite{GANsFormer}, and StytleGAN2~\cite{StyleGAN2}) are released. We plan to include them in our evaluation for future work. Moreover, our work is focused on detecting deepfake and GAN images of humans. Therefore, we have not considered other types of GAN images in this work. Also, we plan to extend our method to detecting full-body deepfakes in the future.

\section{Conclusion}
\label{sec:Conclusion}

Simultaneously detecting a wide range of deepfakes videos and GAN images from different generation methods is a challenging problem. We find that models trained on one type of deepfake (or GAN image) dataset do not perform well on deepfakes (or GAN images) from other generation methods. Therefore, in this work, we formulate and proposed a continual learning approach to learn these new data distributions (or generation method) as new tasks using a small number of data samples in a teacher-student architecture. We also propose three types of loss functions in our objective functions and demonstrate that our CoReD method can effectively mitigate catastrophic forgetting, outperforming other approaches. 

\begin{acks}
This work was partly supported by Institute of Information \& communications Technology Planning \& Evaluation (IITP) grant funded by the Korea government (MSIT) (No.2019-0-00421, AI Graduate School Support Program (Sungkyunkwan University)), (No. 2019-0-01343, Regional strategic industry convergence security core talent training business) and the Basic Science Research Program through National Research Foundation of Korea (NRF) grant funded by Korea government MSIT (No. 2020R1C1C1006004). Also, this research was partly supported by IITP grant funded by the Korea government MSIT (No. 2021-0-00017, Original Technology Development of Artificial Intelligence Industry) and was partly supported by the Korea government MSIT, under the High-Potential Individuals Global Training Program (2020-0-01550) supervised by the IITP. 
\end{acks}

\bibliographystyle{ACM-Reference-Format}
\balance
\bibliography{references}

%%% -*-BibTeX-*-
%%% Do NOT edit. File created by BibTeX with style
%%% ACM-Reference-Format-Journals [18-Jan-2012].

\begin{thebibliography}{80}

%%% ====================================================================
%%% NOTE TO THE USER: you can override these defaults by providing
%%% customized versions of any of these macros before the \bibliography
%%% command.  Each of them MUST provide its own final punctuation,
%%% except for \shownote{}, \showDOI{}, and \showURL{}.  The latter two
%%% do not use final punctuation, in order to avoid confusing it with
%%% the Web address.
%%%
%%% To suppress output of a particular field, define its macro to expand
%%% to an empty string, or better, \unskip, like this:
%%%
%%% \newcommand{\showDOI}[1]{\unskip}   % LaTeX syntax
%%%
%%% \def \showDOI #1{\unskip}           % plain TeX syntax
%%%
%%% ====================================================================

\ifx \showCODEN    \undefined \def \showCODEN     #1{\unskip}     \fi
\ifx \showDOI      \undefined \def \showDOI       #1{#1}\fi
\ifx \showISBNx    \undefined \def \showISBNx     #1{\unskip}     \fi
\ifx \showISBNxiii \undefined \def \showISBNxiii  #1{\unskip}     \fi
\ifx \showISSN     \undefined \def \showISSN      #1{\unskip}     \fi
\ifx \showLCCN     \undefined \def \showLCCN      #1{\unskip}     \fi
\ifx \shownote     \undefined \def \shownote      #1{#1}          \fi
\ifx \showarticletitle \undefined \def \showarticletitle #1{#1}   \fi
\ifx \showURL      \undefined \def \showURL       {\relax}        \fi
% The following commands are used for tagged output and should be
% invisible to TeX
\providecommand\bibfield[2]{#2}
\providecommand\bibinfo[2]{#2}
\providecommand\natexlab[1]{#1}
\providecommand\showeprint[2][]{arXiv:#2}

\bibitem[\protect\citeauthoryear{Afchar, Nozick, Yamagishi, and Echizen}{Afchar
  et~al\mbox{.}}{2018}]%
        {MesoNet}
\bibfield{author}{\bibinfo{person}{Darius Afchar}, \bibinfo{person}{Vincent
  Nozick}, \bibinfo{person}{Junichi Yamagishi}, {and} \bibinfo{person}{Isao
  Echizen}.} \bibinfo{year}{2018}\natexlab{}.
\newblock \showarticletitle{Mesonet: a compact facial video forgery detection
  network}. In \bibinfo{booktitle}{\emph{2018 IEEE International Workshop on
  Information Forensics and Security (WIFS)}}. IEEE, \bibinfo{pages}{1--7}.
\newblock


\bibitem[\protect\citeauthoryear{Agarwal, Farid, El-Gaaly, and Lim}{Agarwal
  et~al\mbox{.}}{2020}]%
        {DeepfakeDetection12}
\bibfield{author}{\bibinfo{person}{Shruti Agarwal}, \bibinfo{person}{Hany
  Farid}, \bibinfo{person}{Tarek El-Gaaly}, {and} \bibinfo{person}{Ser-Nam
  Lim}.} \bibinfo{year}{2020}\natexlab{}.
\newblock \showarticletitle{Detecting deep-fake videos from appearance and
  behavior}. In \bibinfo{booktitle}{\emph{2020 IEEE International Workshop on
  Information Forensics and Security (WIFS)}}. IEEE, \bibinfo{pages}{1--6}.
\newblock


\bibitem[\protect\citeauthoryear{Agarwal, Farid, Gu, He, Nagano, and
  Li}{Agarwal et~al\mbox{.}}{2019}]%
        {DeepfakeDetection11}
\bibfield{author}{\bibinfo{person}{Shruti Agarwal}, \bibinfo{person}{Hany
  Farid}, \bibinfo{person}{Yuming Gu}, \bibinfo{person}{Mingming He},
  \bibinfo{person}{Koki Nagano}, {and} \bibinfo{person}{Hao Li}.}
  \bibinfo{year}{2019}\natexlab{}.
\newblock \showarticletitle{Protecting World Leaders Against Deep Fakes.}. In
  \bibinfo{booktitle}{\emph{CVPR Workshops}}. \bibinfo{pages}{38--45}.
\newblock


\bibitem[\protect\citeauthoryear{Antipov, Baccouche, and Dugelay}{Antipov
  et~al\mbox{.}}{2017}]%
        {FF7}
\bibfield{author}{\bibinfo{person}{Grigory Antipov}, \bibinfo{person}{Moez
  Baccouche}, {and} \bibinfo{person}{Jean-Luc Dugelay}.}
  \bibinfo{year}{2017}\natexlab{}.
\newblock \showarticletitle{Face aging with conditional generative adversarial
  networks}. In \bibinfo{booktitle}{\emph{2017 IEEE International Conference on
  Image Processing (ICIP)}}. IEEE, \bibinfo{pages}{2089--2093}.
\newblock


\bibitem[\protect\citeauthoryear{Bappy, Roy-Chowdhury, Bunk, Nataraj, and
  Manjunath}{Bappy et~al\mbox{.}}{2017}]%
        {DeepfakeDetection4}
\bibfield{author}{\bibinfo{person}{Jawadul~H Bappy}, \bibinfo{person}{Amit~K
  Roy-Chowdhury}, \bibinfo{person}{Jason Bunk}, \bibinfo{person}{Lakshmanan
  Nataraj}, {and} \bibinfo{person}{BS Manjunath}.}
  \bibinfo{year}{2017}\natexlab{}.
\newblock \showarticletitle{Exploiting spatial structure for localizing
  manipulated image regions}. In \bibinfo{booktitle}{\emph{Proceedings of the
  IEEE international conference on computer vision}}.
  \bibinfo{pages}{4970--4979}.
\newblock


\bibitem[\protect\citeauthoryear{Bappy, Simons, Nataraj, Manjunath, and
  Roy-Chowdhury}{Bappy et~al\mbox{.}}{2019}]%
        {DeepfakeDetection3}
\bibfield{author}{\bibinfo{person}{Jawadul~H Bappy}, \bibinfo{person}{Cody
  Simons}, \bibinfo{person}{Lakshmanan Nataraj}, \bibinfo{person}{BS
  Manjunath}, {and} \bibinfo{person}{Amit~K Roy-Chowdhury}.}
  \bibinfo{year}{2019}\natexlab{}.
\newblock \showarticletitle{Hybrid LSTM and encoder--decoder architecture for
  detection of image forgeries}.
\newblock \bibinfo{journal}{\emph{IEEE Transactions on Image Processing}}
  \bibinfo{volume}{28}, \bibinfo{number}{7} (\bibinfo{year}{2019}),
  \bibinfo{pages}{3286--3300}.
\newblock


\bibitem[\protect\citeauthoryear{Chen, Wang, Fu, Long, and Wang}{Chen
  et~al\mbox{.}}{2019}]%
        {chen2019catastrophic}
\bibfield{author}{\bibinfo{person}{Xinyang Chen}, \bibinfo{person}{Sinan Wang},
  \bibinfo{person}{Bo Fu}, \bibinfo{person}{Mingsheng Long}, {and}
  \bibinfo{person}{Jianmin Wang}.} \bibinfo{year}{2019}\natexlab{}.
\newblock \showarticletitle{Catastrophic forgetting meets negative transfer:
  Batch spectral shrinkage for safe transfer learning}.
\newblock  (\bibinfo{year}{2019}).
\newblock


\bibitem[\protect\citeauthoryear{Chen and Liu}{Chen and Liu}{2018}]%
        {CL_Catastrophic1}
\bibfield{author}{\bibinfo{person}{Zhiyuan Chen} {and} \bibinfo{person}{Bing
  Liu}.} \bibinfo{year}{2018}\natexlab{}.
\newblock \showarticletitle{Lifelong machine learning}.
\newblock \bibinfo{journal}{\emph{Synthesis Lectures on Artificial Intelligence
  and Machine Learning}} \bibinfo{volume}{12}, \bibinfo{number}{3}
  (\bibinfo{year}{2018}), \bibinfo{pages}{1--207}.
\newblock


\bibitem[\protect\citeauthoryear{Choi, Choi, Kim, Ha, Kim, and Choo}{Choi
  et~al\mbox{.}}{2018}]%
        {StarGAN}
\bibfield{author}{\bibinfo{person}{Yunjey Choi}, \bibinfo{person}{Minje Choi},
  \bibinfo{person}{Munyoung Kim}, \bibinfo{person}{Jung-Woo Ha},
  \bibinfo{person}{Sunghun Kim}, {and} \bibinfo{person}{Jaegul Choo}.}
  \bibinfo{year}{2018}\natexlab{}.
\newblock \showarticletitle{Stargan: Unified generative adversarial networks
  for multi-domain image-to-image translation}. In
  \bibinfo{booktitle}{\emph{Proceedings of the IEEE Conference on Computer
  Vision and Pattern Recognition}}. \bibinfo{pages}{8789--8797}.
\newblock


\bibitem[\protect\citeauthoryear{Chollet}{Chollet}{2017}]%
        {chollet2017xception}
\bibfield{author}{\bibinfo{person}{Fran{\c{c}}ois Chollet}.}
  \bibinfo{year}{2017}\natexlab{}.
\newblock \showarticletitle{Xception: Deep learning with depthwise separable
  convolutions}. In \bibinfo{booktitle}{\emph{Proceedings of the IEEE
  conference on computer vision and pattern recognition}}.
  \bibinfo{pages}{1251--1258}.
\newblock


\bibitem[\protect\citeauthoryear{Clark, Donahue, and Simonyan}{Clark
  et~al\mbox{.}}{2019}]%
        {VideoGAN_DVDGAN}
\bibfield{author}{\bibinfo{person}{Aidan Clark}, \bibinfo{person}{Jeff
  Donahue}, {and} \bibinfo{person}{Karen Simonyan}.}
  \bibinfo{year}{2019}\natexlab{}.
\newblock \showarticletitle{Efficient video generation on complex datasets}.
\newblock  (\bibinfo{year}{2019}).
\newblock


\bibitem[\protect\citeauthoryear{Cozzolino, Thies, R{\"o}ssler, Riess,
  Nie{\ss}ner, and Verdoliva}{Cozzolino et~al\mbox{.}}{2018}]%
        {cozzolino2018forensictransfer}
\bibfield{author}{\bibinfo{person}{Davide Cozzolino}, \bibinfo{person}{Justus
  Thies}, \bibinfo{person}{Andreas R{\"o}ssler}, \bibinfo{person}{Christian
  Riess}, \bibinfo{person}{Matthias Nie{\ss}ner}, {and} \bibinfo{person}{Luisa
  Verdoliva}.} \bibinfo{year}{2018}\natexlab{}.
\newblock \showarticletitle{Forensictransfer: Weakly-supervised domain
  adaptation for forgery detection}.
\newblock \bibinfo{journal}{\emph{arXiv preprint arXiv:1812.02510}}
  (\bibinfo{year}{2018}).
\newblock


\bibitem[\protect\citeauthoryear{Croft}{Croft}{2019}]%
        {news1}
\bibfield{author}{\bibinfo{person}{Adrian Croft}.}
  \bibinfo{year}{2019}\natexlab{}.
\newblock \bibinfo{title}{From Porn to Scams, Deepfakes Are Becoming a Big
  Racket-And That's Unnerving Business Leaders and Lawmakers}.
\newblock
  \bibinfo{howpublished}{\url{https://fortune.com/2019/10/07/porn-to-scams-deepfakes-big-racket-unnerving-business-leaders-and-lawmakers}}.
\newblock
\newblock
\shownote{[Online; accessed April 16, 2021].}


\bibitem[\protect\citeauthoryear{{Deepfakes Reddit}}{{Deepfakes
  Reddit}}{2018}]%
        {FakeApp}
\bibfield{author}{\bibinfo{person}{{Deepfakes Reddit}}.}
  \bibinfo{year}{2018}\natexlab{}.
\newblock \bibinfo{title}{{FakeApp}}.
\newblock
  \bibinfo{howpublished}{\url{https://www.malavida.com/en/soft/fakeapp}}.
\newblock
\newblock
\shownote{[Online; accessed April 16, 2021].}


\bibitem[\protect\citeauthoryear{Delange, Aljundi, Masana, Parisot, Jia,
  Leonardis, Slabaugh, and Tuytelaars}{Delange et~al\mbox{.}}{2021}]%
        {CoL_Survey}
\bibfield{author}{\bibinfo{person}{Matthias Delange}, \bibinfo{person}{Rahaf
  Aljundi}, \bibinfo{person}{Marc Masana}, \bibinfo{person}{Sarah Parisot},
  \bibinfo{person}{Xu Jia}, \bibinfo{person}{Ales Leonardis},
  \bibinfo{person}{Greg Slabaugh}, {and} \bibinfo{person}{Tinne Tuytelaars}.}
  \bibinfo{year}{2021}\natexlab{}.
\newblock \showarticletitle{A continual learning survey: Defying forgetting in
  classification tasks}.
\newblock \bibinfo{journal}{\emph{IEEE Transactions on Pattern Analysis and
  Machine Intelligence}} (\bibinfo{year}{2021}).
\newblock


\bibitem[\protect\citeauthoryear{Dickson}{Dickson}{2019}]%
        {news2}
\bibfield{author}{\bibinfo{person}{EJ Dickson}.}
  \bibinfo{year}{2019}\natexlab{}.
\newblock \bibinfo{title}{Deepfake Porn Is Still a Threat, Particularly for
  K-Pop Stars}.
\newblock
  \bibinfo{howpublished}{\url{https://www.rollingstone.com/culture/culture-news/deepfakes-nonconsensual-porn-study-kpop-895605}}.
\newblock
\newblock
\shownote{[Online; accessed April 16, 2021].}


\bibitem[\protect\citeauthoryear{Dolhansky, Howes, Pflaum, Baram, and
  Ferrer}{Dolhansky et~al\mbox{.}}{2019}]%
        {DFDC}
\bibfield{author}{\bibinfo{person}{Brian Dolhansky}, \bibinfo{person}{Russ
  Howes}, \bibinfo{person}{Ben Pflaum}, \bibinfo{person}{Nicole Baram}, {and}
  \bibinfo{person}{Cristian~Canton Ferrer}.} \bibinfo{year}{2019}\natexlab{}.
\newblock \showarticletitle{The Deepfake Detection Challenge (DFDC) Preview
  Dataset}.
\newblock \bibinfo{journal}{\emph{arXiv preprint arXiv:1910.08854}}
  (\bibinfo{year}{2019}).
\newblock


\bibitem[\protect\citeauthoryear{Edwards}{Edwards}{2019}]%
        {news4}
\bibfield{author}{\bibinfo{person}{Charlotte Edwards}.}
  \bibinfo{year}{2019}\natexlab{}.
\newblock \bibinfo{title}{Making deepfake porn could soon be as easy as using
  Instagram filters, according to expert}.
\newblock
  \bibinfo{howpublished}{\url{https://www.thesun.co.uk/tech/9800017/deepfake-porn-soon-easy}}.
\newblock
\newblock
\shownote{[Online; accessed April 16, 2021].}


\bibitem[\protect\citeauthoryear{FaceSwapDevs}{FaceSwapDevs}{2019}]%
        {Deepfakes}
\bibfield{author}{\bibinfo{person}{FaceSwapDevs}.}
  \bibinfo{year}{2019}\natexlab{}.
\newblock \bibinfo{title}{Deepfakes\_faceswap - GitHub Repository}.
\newblock \bibinfo{howpublished}{\url{https://github.com/deepfakes/faceswap}}.
\newblock
\newblock
\shownote{[Online; accessed April 16, 2021].}


\bibitem[\protect\citeauthoryear{Goncharov}{Goncharov}{2019}]%
        {FaceApp}
\bibfield{author}{\bibinfo{person}{Yaroslav Goncharov}.}
  \bibinfo{year}{2019}\natexlab{}.
\newblock \bibinfo{title}{{FaceApp - Most Popular Selfie Editor}}.
\newblock \bibinfo{howpublished}{\url{www.faceapp.com}}.
\newblock
\newblock
\shownote{[Online; accessed April 16, 2021].}


\bibitem[\protect\citeauthoryear{G{\"u}era and Delp}{G{\"u}era and
  Delp}{2018}]%
        {FT_SQ2}
\bibfield{author}{\bibinfo{person}{David G{\"u}era} {and}
  \bibinfo{person}{Edward~J Delp}.} \bibinfo{year}{2018}\natexlab{}.
\newblock \showarticletitle{Deepfake video detection using recurrent neural
  networks}. In \bibinfo{booktitle}{\emph{2018 15th IEEE International
  Conference on Advanced Video and Signal Based Surveillance (AVSS)}}. IEEE,
  \bibinfo{pages}{1--6}.
\newblock


\bibitem[\protect\citeauthoryear{Hinton, Vinyals, and Dean}{Hinton
  et~al\mbox{.}}{2015}]%
        {hinton2015distilling}
\bibfield{author}{\bibinfo{person}{Geoffrey Hinton}, \bibinfo{person}{Oriol
  Vinyals}, {and} \bibinfo{person}{Jeff Dean}.}
  \bibinfo{year}{2015}\natexlab{}.
\newblock \showarticletitle{Distilling the knowledge in a neural network}.
\newblock \bibinfo{journal}{\emph{arXiv preprint arXiv:1503.02531}}
  (\bibinfo{year}{2015}).
\newblock


\bibitem[\protect\citeauthoryear{Hou, Pan, Loy, Wang, and Lin}{Hou
  et~al\mbox{.}}{2018}]%
        {COL_KD1}
\bibfield{author}{\bibinfo{person}{Saihui Hou}, \bibinfo{person}{Xinyu Pan},
  \bibinfo{person}{Chen~Change Loy}, \bibinfo{person}{Zilei Wang}, {and}
  \bibinfo{person}{Dahua Lin}.} \bibinfo{year}{2018}\natexlab{}.
\newblock \showarticletitle{Lifelong learning via progressive distillation and
  retrospection}. In \bibinfo{booktitle}{\emph{Proceedings of the European
  Conference on Computer Vision (ECCV)}}. \bibinfo{pages}{437--452}.
\newblock


\bibitem[\protect\citeauthoryear{Huang, Zhang, Li, and He}{Huang
  et~al\mbox{.}}{2017}]%
        {FF34}
\bibfield{author}{\bibinfo{person}{Rui Huang}, \bibinfo{person}{Shu Zhang},
  \bibinfo{person}{Tianyu Li}, {and} \bibinfo{person}{Ran He}.}
  \bibinfo{year}{2017}\natexlab{}.
\newblock \showarticletitle{Beyond face rotation: Global and local perception
  gan for photorealistic and identity preserving frontal view synthesis}. In
  \bibinfo{booktitle}{\emph{Proceedings of the IEEE International Conference on
  Computer Vision}}. \bibinfo{pages}{2439--2448}.
\newblock


\bibitem[\protect\citeauthoryear{Hudson and Zitnick}{Hudson and
  Zitnick}{2021}]%
        {GANsFormer}
\bibfield{author}{\bibinfo{person}{Drew~A Hudson} {and}
  \bibinfo{person}{C~Lawrence Zitnick}.} \bibinfo{year}{2021}\natexlab{}.
\newblock \showarticletitle{Generative Adversarial Transformers}.
\newblock \bibinfo{journal}{\emph{arXiv preprint arXiv:2103.01209}}
  (\bibinfo{year}{2021}).
\newblock


\bibitem[\protect\citeauthoryear{Huh, Liu, Owens, and Efros}{Huh
  et~al\mbox{.}}{2018}]%
        {DeepfakeDetection5}
\bibfield{author}{\bibinfo{person}{Minyoung Huh}, \bibinfo{person}{Andrew Liu},
  \bibinfo{person}{Andrew Owens}, {and} \bibinfo{person}{Alexei~A Efros}.}
  \bibinfo{year}{2018}\natexlab{}.
\newblock \showarticletitle{Fighting fake news: Image splice detection via
  learned self-consistency}. In \bibinfo{booktitle}{\emph{Proceedings of the
  European Conference on Computer Vision (ECCV)}}. \bibinfo{pages}{101--117}.
\newblock


\bibitem[\protect\citeauthoryear{Jeon, Bang, Kim, and Woo}{Jeon
  et~al\mbox{.}}{2020b}]%
        {transferlearning_tgd}
\bibfield{author}{\bibinfo{person}{Hyeonseong Jeon}, \bibinfo{person}{Youngoh
  Bang}, \bibinfo{person}{Junyaup Kim}, {and} \bibinfo{person}{Simon~S Woo}.}
  \bibinfo{year}{2020}\natexlab{b}.
\newblock \showarticletitle{T-GD: Transferable GAN-generated Images Detection
  Framework}.
\newblock \bibinfo{journal}{\emph{arXiv preprint arXiv:2008.04115}}
  (\bibinfo{year}{2020}).
\newblock


\bibitem[\protect\citeauthoryear{Jeon, Bang, and Woo}{Jeon
  et~al\mbox{.}}{2019}]%
        {Hyeonseong2}
\bibfield{author}{\bibinfo{person}{Hyeonseong Jeon}, \bibinfo{person}{Youngoh
  Bang}, {and} \bibinfo{person}{Simon~S Woo}.} \bibinfo{year}{2019}\natexlab{}.
\newblock \showarticletitle{FakeTalkerDetect: Effective and Practical Realistic
  Neural Talking Head Detection with a Highly Unbalanced Dataset}. In
  \bibinfo{booktitle}{\emph{Proceedings of the IEEE International Conference on
  Computer Vision Workshops}}. \bibinfo{pages}{0--0}.
\newblock


\bibitem[\protect\citeauthoryear{Jeon, Bang, and Woo}{Jeon
  et~al\mbox{.}}{2020a}]%
        {Hyeonseong3}
\bibfield{author}{\bibinfo{person}{Hyeonseong Jeon}, \bibinfo{person}{Youngoh
  Bang}, {and} \bibinfo{person}{Simon~S Woo}.}
  \bibinfo{year}{2020}\natexlab{a}.
\newblock \showarticletitle{FDFtNet: Facing Off Fake Images using Fake
  Detection Fine-tuning Network}.
\newblock \bibinfo{journal}{\emph{arXiv preprint arXiv:2001.01265}}
  (\bibinfo{year}{2020}).
\newblock


\bibitem[\protect\citeauthoryear{Kahembwe and Ramamoorthy}{Kahembwe and
  Ramamoorthy}{2020}]%
        {VideoGAN_TGAN-f}
\bibfield{author}{\bibinfo{person}{Emmanuel Kahembwe} {and}
  \bibinfo{person}{Subramanian Ramamoorthy}.} \bibinfo{year}{2020}\natexlab{}.
\newblock \showarticletitle{Lower dimensional kernels for video
  discriminators}.
\newblock \bibinfo{journal}{\emph{Neural Networks}}  \bibinfo{volume}{132}
  (\bibinfo{year}{2020}), \bibinfo{pages}{506--520}.
\newblock


\bibitem[\protect\citeauthoryear{Kan}{Kan}{2019}]%
        {news3}
\bibfield{author}{\bibinfo{person}{Michael Kan}.}
  \bibinfo{year}{2019}\natexlab{}.
\newblock \bibinfo{title}{Most AI-Generated Deepfake Videos Online Are Porn}.
\newblock
  \bibinfo{howpublished}{\url{https://www.pcmag.com/news/371193/most-ai-generated-deepfake-videos-online-are-porn}}.
\newblock
\newblock
\shownote{[Online; accessed April 16, 2021.}


\bibitem[\protect\citeauthoryear{Karras, Aila, Laine, and Lehtinen}{Karras
  et~al\mbox{.}}{2017}]%
        {PGGAN}
\bibfield{author}{\bibinfo{person}{Tero Karras}, \bibinfo{person}{Timo Aila},
  \bibinfo{person}{Samuli Laine}, {and} \bibinfo{person}{Jaakko Lehtinen}.}
  \bibinfo{year}{2017}\natexlab{}.
\newblock \showarticletitle{Progressive growing of gans for improved quality,
  stability, and variation}.
\newblock \bibinfo{journal}{\emph{arXiv preprint arXiv:1710.10196}}
  (\bibinfo{year}{2017}).
\newblock


\bibitem[\protect\citeauthoryear{Karras, Laine, and Aila}{Karras
  et~al\mbox{.}}{2019}]%
        {StyleGAN}
\bibfield{author}{\bibinfo{person}{Tero Karras}, \bibinfo{person}{Samuli
  Laine}, {and} \bibinfo{person}{Timo Aila}.} \bibinfo{year}{2019}\natexlab{}.
\newblock \showarticletitle{A style-based generator architecture for generative
  adversarial networks}. In \bibinfo{booktitle}{\emph{Proceedings of the IEEE
  Conference on Computer Vision and Pattern Recognition}}.
  \bibinfo{pages}{4401--4410}.
\newblock


\bibitem[\protect\citeauthoryear{Karras, Laine, Aittala, Hellsten, Lehtinen,
  and Aila}{Karras et~al\mbox{.}}{2020}]%
        {StyleGAN2}
\bibfield{author}{\bibinfo{person}{Tero Karras}, \bibinfo{person}{Samuli
  Laine}, \bibinfo{person}{Miika Aittala}, \bibinfo{person}{Janne Hellsten},
  \bibinfo{person}{Jaakko Lehtinen}, {and} \bibinfo{person}{Timo Aila}.}
  \bibinfo{year}{2020}\natexlab{}.
\newblock \showarticletitle{Analyzing and improving the image quality of
  stylegan}. In \bibinfo{booktitle}{\emph{Proceedings of the IEEE/CVF
  Conference on Computer Vision and Pattern Recognition}}.
  \bibinfo{pages}{8110--8119}.
\newblock


\bibitem[\protect\citeauthoryear{Kemker, McClure, Abitino, Hayes, and
  Kanan}{Kemker et~al\mbox{.}}{2018}]%
        {catastrophic1}
\bibfield{author}{\bibinfo{person}{Ronald Kemker}, \bibinfo{person}{Marc
  McClure}, \bibinfo{person}{Angelina Abitino}, \bibinfo{person}{Tyler~L
  Hayes}, {and} \bibinfo{person}{Christopher Kanan}.}
  \bibinfo{year}{2018}\natexlab{}.
\newblock \showarticletitle{Measuring catastrophic forgetting in neural
  networks}. In \bibinfo{booktitle}{\emph{Thirty-second AAAI conference on
  artificial intelligence}}.
\newblock


\bibitem[\protect\citeauthoryear{Khalid and Woo}{Khalid and Woo}{2020}]%
        {HASAMCVPRW}
\bibfield{author}{\bibinfo{person}{Hasam Khalid} {and} \bibinfo{person}{Simon~S
  Woo}.} \bibinfo{year}{2020}\natexlab{}.
\newblock \showarticletitle{OC-FakeDect: Classifying deepfakes using one-class
  variational autoencoder}. In \bibinfo{booktitle}{\emph{Proceedings of the
  IEEE/CVF Conference on Computer Vision and Pattern Recognition Workshops}}.
  \bibinfo{pages}{656--657}.
\newblock


\bibitem[\protect\citeauthoryear{Kim, Baddar, Jang, and Ro}{Kim
  et~al\mbox{.}}{2017}]%
        {re2}
\bibfield{author}{\bibinfo{person}{Dae~Hoe Kim}, \bibinfo{person}{Wissam~J
  Baddar}, \bibinfo{person}{Jinhyeok Jang}, {and} \bibinfo{person}{Yong~Man
  Ro}.} \bibinfo{year}{2017}\natexlab{}.
\newblock \showarticletitle{Multi-objective based spatio-temporal feature
  representation learning robust to expression intensity variations for facial
  expression recognition}.
\newblock \bibinfo{journal}{\emph{IEEE Transactions on Affective Computing}}
  \bibinfo{volume}{10}, \bibinfo{number}{2} (\bibinfo{year}{2017}),
  \bibinfo{pages}{223--236}.
\newblock


\bibitem[\protect\citeauthoryear{Kim, Tariq, and Woo}{Kim
  et~al\mbox{.}}{2021}]%
        {Minha_FReTAL}
\bibfield{author}{\bibinfo{person}{Minha Kim}, \bibinfo{person}{Shahroz Tariq},
  {and} \bibinfo{person}{Simon~S Woo}.} \bibinfo{year}{2021}\natexlab{}.
\newblock \showarticletitle{FReTAL: Generalizing Deepfake Detection using
  Knowledge Distillation and Representation Learning}. In
  \bibinfo{booktitle}{\emph{Proceedings of the IEEE/CVF Conference on Computer
  Vision and Pattern Recognition}}. \bibinfo{pages}{1001--1012}.
\newblock


\bibitem[\protect\citeauthoryear{Kirkpatrick, Pascanu, Rabinowitz, Veness,
  Desjardins, Rusu, Milan, Quan, Ramalho, Grabska-Barwinska,
  et~al\mbox{.}}{Kirkpatrick et~al\mbox{.}}{2017}]%
        {catastrophic2}
\bibfield{author}{\bibinfo{person}{James Kirkpatrick}, \bibinfo{person}{Razvan
  Pascanu}, \bibinfo{person}{Neil Rabinowitz}, \bibinfo{person}{Joel Veness},
  \bibinfo{person}{Guillaume Desjardins}, \bibinfo{person}{Andrei~A Rusu},
  \bibinfo{person}{Kieran Milan}, \bibinfo{person}{John Quan},
  \bibinfo{person}{Tiago Ramalho}, \bibinfo{person}{Agnieszka
  Grabska-Barwinska}, {et~al\mbox{.}}} \bibinfo{year}{2017}\natexlab{}.
\newblock \showarticletitle{Overcoming catastrophic forgetting in neural
  networks}.
\newblock \bibinfo{journal}{\emph{Proceedings of the national academy of
  sciences}} \bibinfo{volume}{114}, \bibinfo{number}{13}
  (\bibinfo{year}{2017}), \bibinfo{pages}{3521--3526}.
\newblock


\bibitem[\protect\citeauthoryear{Kowalski}{Kowalski}{2016}]%
        {FaceSwap}
\bibfield{author}{\bibinfo{person}{Marek Kowalski}.}
  \bibinfo{year}{2016}\natexlab{}.
\newblock \bibinfo{title}{FaceSwap - GitHub Repository}.
\newblock
  \bibinfo{howpublished}{\url{https://github.com/MarekKowalski/FaceSwap}}.
\newblock
\newblock
\shownote{[Online; accessed April 16, 2021].}


\bibitem[\protect\citeauthoryear{Lample, Zeghidour, Usunier, Bordes, Denoyer,
  and Ranzato}{Lample et~al\mbox{.}}{2017}]%
        {FF43}
\bibfield{author}{\bibinfo{person}{Guillaume Lample}, \bibinfo{person}{Neil
  Zeghidour}, \bibinfo{person}{Nicolas Usunier}, \bibinfo{person}{Antoine
  Bordes}, \bibinfo{person}{Ludovic Denoyer}, {and}
  \bibinfo{person}{Marc'Aurelio Ranzato}.} \bibinfo{year}{2017}\natexlab{}.
\newblock \showarticletitle{Fader networks: Manipulating images by sliding
  attributes}. In \bibinfo{booktitle}{\emph{Advances in Neural Information
  Processing Systems}}. \bibinfo{pages}{5967--5976}.
\newblock


\bibitem[\protect\citeauthoryear{Lee, Tariq, Kim, and Woo}{Lee
  et~al\mbox{.}}{2021a}]%
        {SAM_TAR}
\bibfield{author}{\bibinfo{person}{Sangyup Lee}, \bibinfo{person}{Shahroz
  Tariq}, \bibinfo{person}{Junyaup Kim}, {and} \bibinfo{person}{Simon~S Woo}.}
  \bibinfo{year}{2021}\natexlab{a}.
\newblock \showarticletitle{TAR: Generalized Forensic Framework to Detect
  Deepfakes Using Weakly Supervised Learning}. In
  \bibinfo{booktitle}{\emph{IFIP International Conference on ICT Systems
  Security and Privacy Protection}}. Springer, \bibinfo{pages}{351--366}.
\newblock


\bibitem[\protect\citeauthoryear{Lee, Tariq, Shin, and Woo}{Lee
  et~al\mbox{.}}{2021b}]%
        {SAMGAN}
\bibfield{author}{\bibinfo{person}{Sangyup Lee}, \bibinfo{person}{Shahroz
  Tariq}, \bibinfo{person}{Youjin Shin}, {and} \bibinfo{person}{Simon~S Woo}.}
  \bibinfo{year}{2021}\natexlab{b}.
\newblock \showarticletitle{Detecting handcrafted facial image manipulations
  and GAN-generated facial images using Shallow-FakeFaceNet}.
\newblock \bibinfo{journal}{\emph{Applied Soft Computing}}
  \bibinfo{volume}{105} (\bibinfo{year}{2021}), \bibinfo{pages}{107256}.
\newblock


\bibitem[\protect\citeauthoryear{Li, Bao, Zhang, Yang, Chen, Wen, and Guo}{Li
  et~al\mbox{.}}{2020a}]%
        {FaceXRay}
\bibfield{author}{\bibinfo{person}{Lingzhi Li}, \bibinfo{person}{Jianmin Bao},
  \bibinfo{person}{Ting Zhang}, \bibinfo{person}{Hao Yang},
  \bibinfo{person}{Dong Chen}, \bibinfo{person}{Fang Wen}, {and}
  \bibinfo{person}{Baining Guo}.} \bibinfo{year}{2020}\natexlab{a}.
\newblock \showarticletitle{Face x-ray for more general face forgery
  detection}. In \bibinfo{booktitle}{\emph{Proceedings of the IEEE/CVF
  Conference on Computer Vision and Pattern Recognition}}.
  \bibinfo{pages}{5001--5010}.
\newblock


\bibitem[\protect\citeauthoryear{Li, Yang, Sun, Qi, and Lyu}{Li
  et~al\mbox{.}}{2020b}]%
        {CelebDF}
\bibfield{author}{\bibinfo{person}{Yuezun Li}, \bibinfo{person}{Xin Yang},
  \bibinfo{person}{Pu Sun}, \bibinfo{person}{Honggang Qi}, {and}
  \bibinfo{person}{Siwei Lyu}.} \bibinfo{year}{2020}\natexlab{b}.
\newblock \showarticletitle{Celeb-df: A large-scale challenging dataset for
  deepfake forensics}. In \bibinfo{booktitle}{\emph{Proceedings of the IEEE/CVF
  Conference on Computer Vision and Pattern Recognition}}.
  \bibinfo{pages}{3207--3216}.
\newblock


\bibitem[\protect\citeauthoryear{Li and Hoiem}{Li and Hoiem}{2017}]%
        {li2017learning}
\bibfield{author}{\bibinfo{person}{Zhizhong Li} {and} \bibinfo{person}{Derek
  Hoiem}.} \bibinfo{year}{2017}\natexlab{}.
\newblock \showarticletitle{Learning without forgetting}.
\newblock \bibinfo{journal}{\emph{IEEE transactions on pattern analysis and
  machine intelligence}} \bibinfo{volume}{40}, \bibinfo{number}{12}
  (\bibinfo{year}{2017}), \bibinfo{pages}{2935--2947}.
\newblock


\bibitem[\protect\citeauthoryear{Liu, Luo, Wang, and Tang}{Liu
  et~al\mbox{.}}{2015}]%
        {CelebA}
\bibfield{author}{\bibinfo{person}{Ziwei Liu}, \bibinfo{person}{Ping Luo},
  \bibinfo{person}{Xiaogang Wang}, {and} \bibinfo{person}{Xiaoou Tang}.}
  \bibinfo{year}{2015}\natexlab{}.
\newblock \showarticletitle{Deep Learning Face Attributes in the Wild}. In
  \bibinfo{booktitle}{\emph{Proceedings of International Conference on Computer
  Vision (ICCV)}}.
\newblock


\bibitem[\protect\citeauthoryear{Long, Cao, Cao, Wang, and Jordan}{Long
  et~al\mbox{.}}{2018}]%
        {re1}
\bibfield{author}{\bibinfo{person}{Mingsheng Long}, \bibinfo{person}{Yue Cao},
  \bibinfo{person}{Zhangjie Cao}, \bibinfo{person}{Jianmin Wang}, {and}
  \bibinfo{person}{Michael~I Jordan}.} \bibinfo{year}{2018}\natexlab{}.
\newblock \showarticletitle{Transferable representation learning with deep
  adaptation networks}.
\newblock \bibinfo{journal}{\emph{IEEE transactions on pattern analysis and
  machine intelligence}} \bibinfo{volume}{41}, \bibinfo{number}{12}
  (\bibinfo{year}{2018}), \bibinfo{pages}{3071--3085}.
\newblock


\bibitem[\protect\citeauthoryear{Lu, Tai, and Tang}{Lu et~al\mbox{.}}{2017b}]%
        {FF46}
\bibfield{author}{\bibinfo{person}{Yongyi Lu}, \bibinfo{person}{Yu-Wing Tai},
  {and} \bibinfo{person}{Chi-Keung Tang}.} \bibinfo{year}{2017}\natexlab{b}.
\newblock \showarticletitle{Conditional cyclegan for attribute guided face
  image generation}.
\newblock \bibinfo{journal}{\emph{arXiv preprint arXiv:1705.09966}}
  (\bibinfo{year}{2017}).
\newblock


\bibitem[\protect\citeauthoryear{Lu, Li, Cao, He, and Sun}{Lu
  et~al\mbox{.}}{2017a}]%
        {SurveyFaceSynthesis}
\bibfield{author}{\bibinfo{person}{Zhihe Lu}, \bibinfo{person}{Zhihang Li},
  \bibinfo{person}{Jie Cao}, \bibinfo{person}{Ran He}, {and}
  \bibinfo{person}{Zhenan Sun}.} \bibinfo{year}{2017}\natexlab{a}.
\newblock \showarticletitle{Recent progress of face image synthesis}. In
  \bibinfo{booktitle}{\emph{2017 4th IAPR Asian Conference on Pattern
  Recognition (ACPR)}}. IEEE, \bibinfo{pages}{7--12}.
\newblock


\bibitem[\protect\citeauthoryear{Marra, Saltori, Boato, and Verdoliva}{Marra
  et~al\mbox{.}}{2019}]%
        {incremental_gan}
\bibfield{author}{\bibinfo{person}{Francesco Marra}, \bibinfo{person}{Cristiano
  Saltori}, \bibinfo{person}{Giulia Boato}, {and} \bibinfo{person}{Luisa
  Verdoliva}.} \bibinfo{year}{2019}\natexlab{}.
\newblock \showarticletitle{Incremental learning for the detection and
  classification of gan-generated images}. In \bibinfo{booktitle}{\emph{2019
  IEEE International Workshop on Information Forensics and Security (WIFS)}}.
  IEEE, \bibinfo{pages}{1--6}.
\newblock


\bibitem[\protect\citeauthoryear{Mehta}{Mehta}{2019}]%
        {news5}
\bibfield{author}{\bibinfo{person}{Ivan Mehta}.}
  \bibinfo{year}{2019}\natexlab{}.
\newblock \bibinfo{title}{A new study says nearly 96 of deepfake videos are
  porn}.
\newblock
  \bibinfo{howpublished}{\url{https://thenextweb.com/apps/2019/10/07/a-new-study-says-nearly-96-of-deepfake-videos-are-porn}}.
\newblock
\newblock
\shownote{Accessed: 2020-02-11.}


\bibitem[\protect\citeauthoryear{Michieli and Zanuttigh}{Michieli and
  Zanuttigh}{2021}]%
        {COL_KD2}
\bibfield{author}{\bibinfo{person}{Umberto Michieli} {and}
  \bibinfo{person}{Pietro Zanuttigh}.} \bibinfo{year}{2021}\natexlab{}.
\newblock \showarticletitle{Knowledge distillation for incremental learning in
  semantic segmentation}.
\newblock \bibinfo{journal}{\emph{Computer Vision and Image Understanding}}
  \bibinfo{volume}{205} (\bibinfo{year}{2021}), \bibinfo{pages}{103167}.
\newblock


\bibitem[\protect\citeauthoryear{Mirsky and Lee}{Mirsky and Lee}{2021}]%
        {WenkeSurvey}
\bibfield{author}{\bibinfo{person}{Yisroel Mirsky} {and} \bibinfo{person}{Wenke
  Lee}.} \bibinfo{year}{2021}\natexlab{}.
\newblock \showarticletitle{The creation and detection of deepfakes: A survey}.
\newblock \bibinfo{journal}{\emph{ACM Computing Surveys (CSUR)}}
  \bibinfo{volume}{54}, \bibinfo{number}{1} (\bibinfo{year}{2021}),
  \bibinfo{pages}{1--41}.
\newblock


\bibitem[\protect\citeauthoryear{{Momo}}{{Momo}}{2020}]%
        {ZAO}
\bibfield{author}{\bibinfo{person}{{Momo}}.} \bibinfo{year}{2020}\natexlab{}.
\newblock \bibinfo{title}{{ZAOAPP}}.
\newblock \bibinfo{howpublished}{\url{https://zao.en.softonic.com/android}}.
\newblock
\newblock
\shownote{[Online; accessed April 16, 2021].}


\bibitem[\protect\citeauthoryear{Parisi, Kemker, Part, Kanan, and
  Wermter}{Parisi et~al\mbox{.}}{2019}]%
        {CoL_Survey2}
\bibfield{author}{\bibinfo{person}{German~I Parisi}, \bibinfo{person}{Ronald
  Kemker}, \bibinfo{person}{Jose~L Part}, \bibinfo{person}{Christopher Kanan},
  {and} \bibinfo{person}{Stefan Wermter}.} \bibinfo{year}{2019}\natexlab{}.
\newblock \showarticletitle{Continual lifelong learning with neural networks: A
  review}.
\newblock \bibinfo{journal}{\emph{Neural Networks}}  \bibinfo{volume}{113}
  (\bibinfo{year}{2019}), \bibinfo{pages}{54--71}.
\newblock


\bibitem[\protect\citeauthoryear{Razavi, Oord, and Vinyals}{Razavi
  et~al\mbox{.}}{2019}]%
        {VQVAE2}
\bibfield{author}{\bibinfo{person}{Ali Razavi}, \bibinfo{person}{Aaron van~den
  Oord}, {and} \bibinfo{person}{Oriol Vinyals}.}
  \bibinfo{year}{2019}\natexlab{}.
\newblock \showarticletitle{Generating Diverse High-Fidelity Images with
  VQ-VAE-2}.
\newblock \bibinfo{journal}{\emph{arXiv preprint arXiv:1906.00446}}
  (\bibinfo{year}{2019}).
\newblock


\bibitem[\protect\citeauthoryear{Rebuffi, Kolesnikov, Sperl, and
  Lampert}{Rebuffi et~al\mbox{.}}{2017}]%
        {rebuffi2017icarl}
\bibfield{author}{\bibinfo{person}{Sylvestre-Alvise Rebuffi},
  \bibinfo{person}{Alexander Kolesnikov}, \bibinfo{person}{Georg Sperl}, {and}
  \bibinfo{person}{Christoph~H Lampert}.} \bibinfo{year}{2017}\natexlab{}.
\newblock \showarticletitle{icarl: Incremental classifier and representation
  learning}. In \bibinfo{booktitle}{\emph{Proceedings of the IEEE conference on
  Computer Vision and Pattern Recognition}}. \bibinfo{pages}{2001--2010}.
\newblock


\bibitem[\protect\citeauthoryear{R\"ossler, Cozzolino, Verdoliva, Riess, Thies,
  and Nie{\ss}ner}{R\"ossler et~al\mbox{.}}{2019}]%
        {FaceForensics++}
\bibfield{author}{\bibinfo{person}{Andreas R\"ossler}, \bibinfo{person}{Davide
  Cozzolino}, \bibinfo{person}{Luisa Verdoliva}, \bibinfo{person}{Christian
  Riess}, \bibinfo{person}{Justus Thies}, {and} \bibinfo{person}{Matthias
  Nie{\ss}ner}.} \bibinfo{year}{2019}\natexlab{}.
\newblock \showarticletitle{FaceForensics++: Learning to Detect Manipulated
  Facial Images}. In \bibinfo{booktitle}{\emph{ICCV 2019}}.
\newblock


\bibitem[\protect\citeauthoryear{Sabir, Cheng, Jaiswal, AbdAlmageed, Masi, and
  Natarajan}{Sabir et~al\mbox{.}}{2019}]%
        {DFD2}
\bibfield{author}{\bibinfo{person}{Ekraam Sabir}, \bibinfo{person}{Jiaxin
  Cheng}, \bibinfo{person}{Ayush Jaiswal}, \bibinfo{person}{Wael AbdAlmageed},
  \bibinfo{person}{Iacopo Masi}, {and} \bibinfo{person}{Prem Natarajan}.}
  \bibinfo{year}{2019}\natexlab{}.
\newblock \showarticletitle{Recurrent Convolutional Strategies for Face
  Manipulation Detection in Videos}.
\newblock \bibinfo{journal}{\emph{Interfaces (GUI)}}  \bibinfo{volume}{3}
  (\bibinfo{year}{2019}), \bibinfo{pages}{1}.
\newblock


\bibitem[\protect\citeauthoryear{Saito, Saito, Koyama, and Kobayashi}{Saito
  et~al\mbox{.}}{2020}]%
        {VideoGAN_TGANv2}
\bibfield{author}{\bibinfo{person}{Masaki Saito}, \bibinfo{person}{Shunta
  Saito}, \bibinfo{person}{Masanori Koyama}, {and} \bibinfo{person}{Sosuke
  Kobayashi}.} \bibinfo{year}{2020}\natexlab{}.
\newblock \showarticletitle{Train Sparsely, Generate Densely: Memory-Efficient
  Unsupervised Training of High-Resolution Temporal GAN}.
\newblock \bibinfo{journal}{\emph{International Journal of Computer Vision}}
  \bibinfo{volume}{128} (\bibinfo{year}{2020}), \bibinfo{pages}{2586--2606}.
\newblock


\bibitem[\protect\citeauthoryear{Salloum, Ren, and Kuo}{Salloum
  et~al\mbox{.}}{2018}]%
        {DeepfakeDetection6}
\bibfield{author}{\bibinfo{person}{Ronald Salloum}, \bibinfo{person}{Yuzhuo
  Ren}, {and} \bibinfo{person}{C-C~Jay Kuo}.} \bibinfo{year}{2018}\natexlab{}.
\newblock \showarticletitle{Image splicing localization using a multi-task
  fully convolutional network (MFCN)}.
\newblock \bibinfo{journal}{\emph{Journal of Visual Communication and Image
  Representation}}  \bibinfo{volume}{51} (\bibinfo{year}{2018}),
  \bibinfo{pages}{201--209}.
\newblock


\bibitem[\protect\citeauthoryear{Tan and Le}{Tan and Le}{2019}]%
        {Efficientnet}
\bibfield{author}{\bibinfo{person}{Mingxing Tan} {and} \bibinfo{person}{Quoc
  Le}.} \bibinfo{year}{2019}\natexlab{}.
\newblock \showarticletitle{Efficientnet: Rethinking model scaling for
  convolutional neural networks}. In \bibinfo{booktitle}{\emph{International
  Conference on Machine Learning}}. PMLR, \bibinfo{pages}{6105--6114}.
\newblock


\bibitem[\protect\citeauthoryear{Tariq, Jeon, and Woo}{Tariq
  et~al\mbox{.}}{2021a}]%
        {ShahrozAPI}
\bibfield{author}{\bibinfo{person}{Shahroz Tariq}, \bibinfo{person}{Sowon
  Jeon}, {and} \bibinfo{person}{Simon~S Woo}.}
  \bibinfo{year}{2021}\natexlab{a}.
\newblock \showarticletitle{Am I a Real or Fake Celebrity? Measuring Commercial
  Face Recognition Web APIs under Deepfake Impersonation Attack}.
\newblock \bibinfo{journal}{\emph{arXiv preprint arXiv:2103.00847}}
  (\bibinfo{year}{2021}).
\newblock


\bibitem[\protect\citeauthoryear{Tariq, Lee, Kim, Shin, and Woo}{Tariq
  et~al\mbox{.}}{2018}]%
        {Shahroz1}
\bibfield{author}{\bibinfo{person}{Shahroz Tariq}, \bibinfo{person}{Sangyup
  Lee}, \bibinfo{person}{Hoyoung Kim}, \bibinfo{person}{Youjin Shin}, {and}
  \bibinfo{person}{Simon~S Woo}.} \bibinfo{year}{2018}\natexlab{}.
\newblock \showarticletitle{Detecting both machine and human created fake face
  images in the wild}. In \bibinfo{booktitle}{\emph{Proceedings of the 2nd
  International Workshop on Multimedia Privacy and Security}}. ACM,
  \bibinfo{pages}{81--87}.
\newblock


\bibitem[\protect\citeauthoryear{Tariq, Lee, Kim, Shin, and Woo}{Tariq
  et~al\mbox{.}}{2019}]%
        {Shahroz2}
\bibfield{author}{\bibinfo{person}{Shahroz Tariq}, \bibinfo{person}{Sangyup
  Lee}, \bibinfo{person}{Hoyoung Kim}, \bibinfo{person}{Youjin Shin}, {and}
  \bibinfo{person}{Simon~S Woo}.} \bibinfo{year}{2019}\natexlab{}.
\newblock \showarticletitle{GAN is a friend or foe?: a framework to detect
  various fake face images}. In \bibinfo{booktitle}{\emph{Proceedings of the
  34th ACM/SIGAPP Symposium on Applied Computing}}. ACM,
  \bibinfo{pages}{1296--1303}.
\newblock


\bibitem[\protect\citeauthoryear{Tariq, Lee, and Woo}{Tariq
  et~al\mbox{.}}{2021b}]%
        {Shahroz3}
\bibfield{author}{\bibinfo{person}{Shahroz Tariq}, \bibinfo{person}{Sangyup
  Lee}, {and} \bibinfo{person}{Simon Woo}.} \bibinfo{year}{2021}\natexlab{b}.
\newblock \showarticletitle{One detector to rule them all: Towards a general
  deepfake attack detection framework}. In
  \bibinfo{booktitle}{\emph{Proceedings of the Web Conference 2021}}.
  \bibinfo{pages}{3625--3637}.
\newblock
\urldef\tempurl%
\url{https://doi.org/10.1145/3442381.3449809}
\showDOI{\tempurl}


\bibitem[\protect\citeauthoryear{Tariq, Lee, and Woo}{Tariq
  et~al\mbox{.}}{2020}]%
        {CLRNet}
\bibfield{author}{\bibinfo{person}{Shahroz Tariq}, \bibinfo{person}{Sangyup
  Lee}, {and} \bibinfo{person}{Simon~S Woo}.} \bibinfo{year}{2020}\natexlab{}.
\newblock \showarticletitle{A Convolutional LSTM based Residual Network for
  Deepfake Video Detection}.
\newblock \bibinfo{journal}{\emph{arXiv preprint arXiv:2009.07480}}
  (\bibinfo{year}{2020}).
\newblock


\bibitem[\protect\citeauthoryear{Thai, Stojanov, Rehg, and Rehg}{Thai
  et~al\mbox{.}}{2021}]%
        {CL_Catastrophic2}
\bibfield{author}{\bibinfo{person}{Anh Thai}, \bibinfo{person}{Stefan
  Stojanov}, \bibinfo{person}{Isaac Rehg}, {and} \bibinfo{person}{James~M
  Rehg}.} \bibinfo{year}{2021}\natexlab{}.
\newblock \showarticletitle{Does Continual Learning= Catastrophic Forgetting?}
\newblock \bibinfo{journal}{\emph{arXiv preprint arXiv:2101.07295}}
  (\bibinfo{year}{2021}).
\newblock


\bibitem[\protect\citeauthoryear{Thies, Zollh{\"o}fer, and Nie{\ss}ner}{Thies
  et~al\mbox{.}}{2019}]%
        {NeuralTextures}
\bibfield{author}{\bibinfo{person}{Justus Thies}, \bibinfo{person}{Michael
  Zollh{\"o}fer}, {and} \bibinfo{person}{Matthias Nie{\ss}ner}.}
  \bibinfo{year}{2019}\natexlab{}.
\newblock \showarticletitle{Deferred neural rendering: Image synthesis using
  neural textures}.
\newblock \bibinfo{journal}{\emph{ACM Transactions on Graphics (TOG)}}
  \bibinfo{volume}{38}, \bibinfo{number}{4} (\bibinfo{year}{2019}),
  \bibinfo{pages}{1--12}.
\newblock


\bibitem[\protect\citeauthoryear{Thies, Zollh\"{o}fer, Stamminger, Theobalt,
  and Nie{\ss}ner}{Thies et~al\mbox{.}}{2018}]%
        {Face2Face}
\bibfield{author}{\bibinfo{person}{Justus Thies}, \bibinfo{person}{Michael
  Zollh\"{o}fer}, \bibinfo{person}{Marc Stamminger}, \bibinfo{person}{Christian
  Theobalt}, {and} \bibinfo{person}{Matthias Nie{\ss}ner}.}
  \bibinfo{year}{2018}\natexlab{}.
\newblock \showarticletitle{Face2Face: Real-time Face Capture and Reenactment
  of RGB Videos}.
\newblock \bibinfo{journal}{\emph{Commun. ACM}} \bibinfo{volume}{62},
  \bibinfo{number}{1} (\bibinfo{date}{Dec.} \bibinfo{year}{2018}),
  \bibinfo{pages}{96--104}.
\newblock
\showISSN{0001-0782}


\bibitem[\protect\citeauthoryear{Thompson, Gwinnup, Khayrallah, Duh, and
  Koehn}{Thompson et~al\mbox{.}}{2019}]%
        {Catastrophic3}
\bibfield{author}{\bibinfo{person}{Brian Thompson}, \bibinfo{person}{Jeremy
  Gwinnup}, \bibinfo{person}{Huda Khayrallah}, \bibinfo{person}{Kevin Duh},
  {and} \bibinfo{person}{Philipp Koehn}.} \bibinfo{year}{2019}\natexlab{}.
\newblock \showarticletitle{Overcoming catastrophic forgetting during domain
  adaptation of neural machine translation}. In
  \bibinfo{booktitle}{\emph{Proceedings of the 2019 Conference of the North
  American Chapter of the Association for Computational Linguistics: Human
  Language Technologies, Volume 1 (Long and Short Papers)}}.
  \bibinfo{pages}{2062--2068}.
\newblock


\bibitem[\protect\citeauthoryear{Upchurch, Gardner, Pleiss, Pless, Snavely,
  Bala, and Weinberger}{Upchurch et~al\mbox{.}}{2017}]%
        {FF62}
\bibfield{author}{\bibinfo{person}{Paul Upchurch}, \bibinfo{person}{Jacob
  Gardner}, \bibinfo{person}{Geoff Pleiss}, \bibinfo{person}{Robert Pless},
  \bibinfo{person}{Noah Snavely}, \bibinfo{person}{Kavita Bala}, {and}
  \bibinfo{person}{Kilian Weinberger}.} \bibinfo{year}{2017}\natexlab{}.
\newblock \showarticletitle{Deep feature interpolation for image content
  changes}. In \bibinfo{booktitle}{\emph{Proceedings of the IEEE conference on
  computer vision and pattern recognition}}. \bibinfo{pages}{7064--7073}.
\newblock


\bibitem[\protect\citeauthoryear{Xu, Zhong, Yepes, and Lau}{Xu
  et~al\mbox{.}}{2020}]%
        {Catastrophic4}
\bibfield{author}{\bibinfo{person}{Ying Xu}, \bibinfo{person}{Xu Zhong},
  \bibinfo{person}{Antonio Jose~Jimeno Yepes}, {and} \bibinfo{person}{Jey~Han
  Lau}.} \bibinfo{year}{2020}\natexlab{}.
\newblock \showarticletitle{Forget me not: Reducing catastrophic forgetting for
  domain adaptation in reading comprehension}. In
  \bibinfo{booktitle}{\emph{2020 International Joint Conference on Neural
  Networks (IJCNN)}}. IEEE, \bibinfo{pages}{1--8}.
\newblock


\bibitem[\protect\citeauthoryear{Yang, Li, and Lyu}{Yang et~al\mbox{.}}{2019}]%
        {DeepfakeDetection10}
\bibfield{author}{\bibinfo{person}{Xin Yang}, \bibinfo{person}{Yuezun Li},
  {and} \bibinfo{person}{Siwei Lyu}.} \bibinfo{year}{2019}\natexlab{}.
\newblock \showarticletitle{Exposing deep fakes using inconsistent head poses}.
  In \bibinfo{booktitle}{\emph{ICASSP 2019-2019 IEEE International Conference
  on Acoustics, Speech and Signal Processing (ICASSP)}}. IEEE,
  \bibinfo{pages}{8261--8265}.
\newblock


\bibitem[\protect\citeauthoryear{Yun, Han, Oh, Chun, Choe, and Yoo}{Yun
  et~al\mbox{.}}{2019}]%
        {yun2019cutmix}
\bibfield{author}{\bibinfo{person}{Sangdoo Yun}, \bibinfo{person}{Dongyoon
  Han}, \bibinfo{person}{Seong~Joon Oh}, \bibinfo{person}{Sanghyuk Chun},
  \bibinfo{person}{Junsuk Choe}, {and} \bibinfo{person}{Youngjoon Yoo}.}
  \bibinfo{year}{2019}\natexlab{}.
\newblock \showarticletitle{Cutmix: Regularization strategy to train strong
  classifiers with localizable features}. In
  \bibinfo{booktitle}{\emph{Proceedings of the IEEE/CVF International
  Conference on Computer Vision}}. \bibinfo{pages}{6023--6032}.
\newblock


\bibitem[\protect\citeauthoryear{Zhang, Zhang, Li, and Qiao}{Zhang
  et~al\mbox{.}}{2016}]%
        {MTCNN}
\bibfield{author}{\bibinfo{person}{Kaipeng Zhang}, \bibinfo{person}{Zhanpeng
  Zhang}, \bibinfo{person}{Zhifeng Li}, {and} \bibinfo{person}{Yu Qiao}.}
  \bibinfo{year}{2016}\natexlab{}.
\newblock \showarticletitle{Joint face detection and alignment using multitask
  cascaded convolutional networks}.
\newblock \bibinfo{journal}{\emph{IEEE Signal Processing Letters}}
  \bibinfo{volume}{23}, \bibinfo{number}{10} (\bibinfo{year}{2016}),
  \bibinfo{pages}{1499--1503}.
\newblock


\bibitem[\protect\citeauthoryear{Zhou, Han, Morariu, and Davis}{Zhou
  et~al\mbox{.}}{2017}]%
        {DeepfakeDetection1}
\bibfield{author}{\bibinfo{person}{Peng Zhou}, \bibinfo{person}{Xintong Han},
  \bibinfo{person}{Vlad~I Morariu}, {and} \bibinfo{person}{Larry~S Davis}.}
  \bibinfo{year}{2017}\natexlab{}.
\newblock \showarticletitle{Two-stream neural networks for tampered face
  detection}. In \bibinfo{booktitle}{\emph{2017 IEEE Conference on Computer
  Vision and Pattern Recognition Workshops (CVPRW)}}. IEEE,
  \bibinfo{pages}{1831--1839}.
\newblock


\bibitem[\protect\citeauthoryear{Zhou, Han, Morariu, and Davis}{Zhou
  et~al\mbox{.}}{2018}]%
        {DeepfakeDetection2}
\bibfield{author}{\bibinfo{person}{Peng Zhou}, \bibinfo{person}{Xintong Han},
  \bibinfo{person}{Vlad~I Morariu}, {and} \bibinfo{person}{Larry~S Davis}.}
  \bibinfo{year}{2018}\natexlab{}.
\newblock \showarticletitle{Learning rich features for image manipulation
  detection}. In \bibinfo{booktitle}{\emph{Proceedings of the IEEE Conference
  on Computer Vision and Pattern Recognition}}. \bibinfo{pages}{1053--1061}.
\newblock


\bibitem[\protect\citeauthoryear{Zi, Chang, Chen, Ma, and Jiang}{Zi
  et~al\mbox{.}}{2020}]%
        {WildDeepfake}
\bibfield{author}{\bibinfo{person}{Bojia Zi}, \bibinfo{person}{Minghao Chang},
  \bibinfo{person}{Jingjing Chen}, \bibinfo{person}{Xingjun Ma}, {and}
  \bibinfo{person}{Yu-Gang Jiang}.} \bibinfo{year}{2020}\natexlab{}.
\newblock \showarticletitle{WildDeepfake: A Challenging Real-World Dataset for
  Deepfake Detection}. In \bibinfo{booktitle}{\emph{Proceedings of the 28th ACM
  International Conference on Multimedia}}. \bibinfo{pages}{2382--2390}.
\newblock


\end{thebibliography}

\clearpage
\onecolumn
\appendix

\section{Supplementary Materials}
\begin{figure*}[htp!]
    % \vspace{30pt}
    \centering
    \includegraphics[width=1\linewidth]{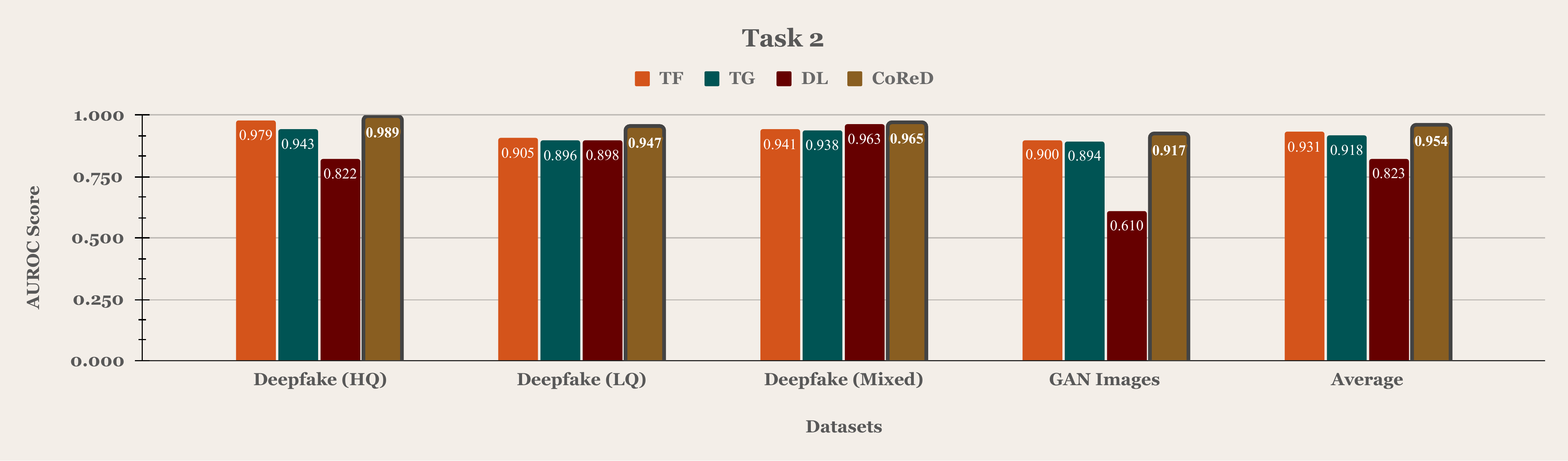}
    \caption{Task 2 results (AUROC) for deepfake video and GAN image detection. All methods perform relatively well. However, CoReD shows the best performance (outlined in black and boldfaced).}
    \label{fig:AUC_TASK2}
    % \vspace{50pt}
\end{figure*}

\begin{figure*}[htp!]
    \centering
    \includegraphics[width=1\linewidth]{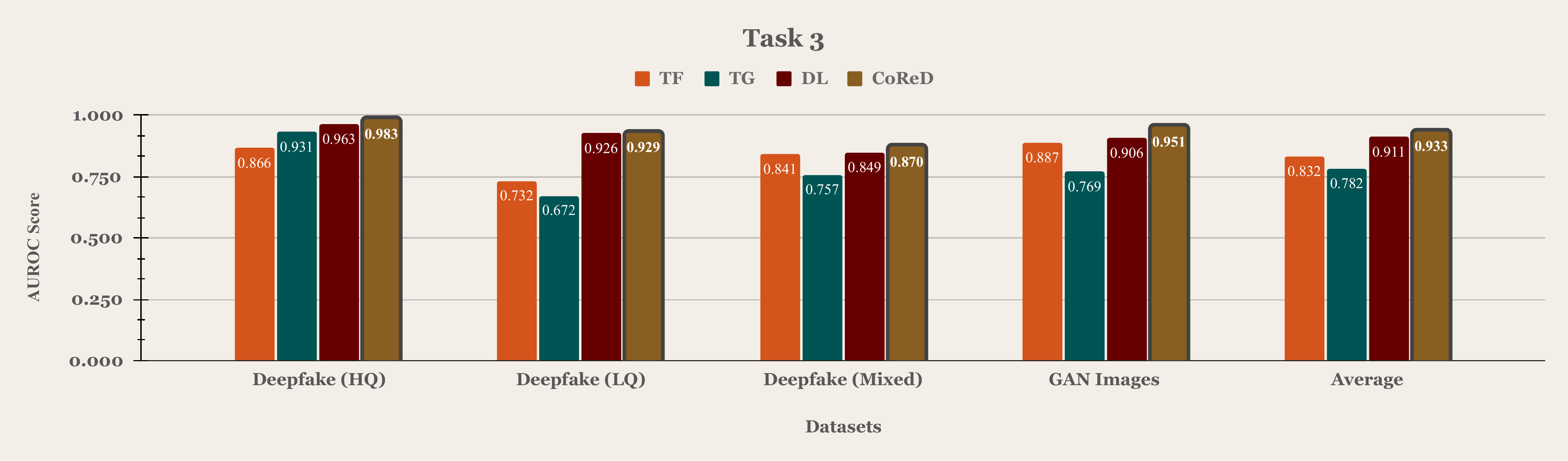}
    \caption{Task 3 results (AUROC) for deepfake video and GAN image detection. TF and TG suffer from catastrophic forgetting, whereas DL and CoReD demonstrate good performance. The best performer is outlined in black and boldfaced.}
    \label{fig:AUC_TASK3}
\end{figure*}

\begin{figure*}[htp!]
    \centering
    \includegraphics[width=1\linewidth]{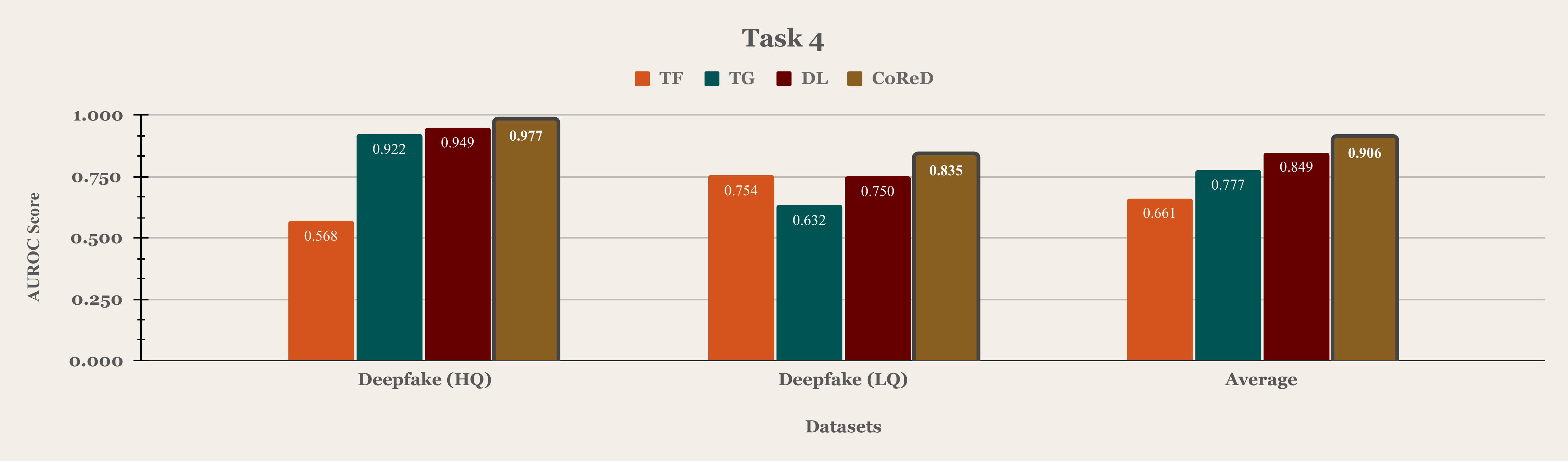}
    \caption{Task 4 results (AUROC) for deepfake video detection. All baseline (i.e., TF, TG, and DL) suffers from catastrophic forgetting, whereas CoReD demonstrates stable and consistent performance (outlined in black and boldfaced).}
    \label{fig:AUC_TASK4}
    % \vspace{25pt}
\end{figure*}

\begin{figure*}[htp!]
    \centering
    \includegraphics[width=1\linewidth]{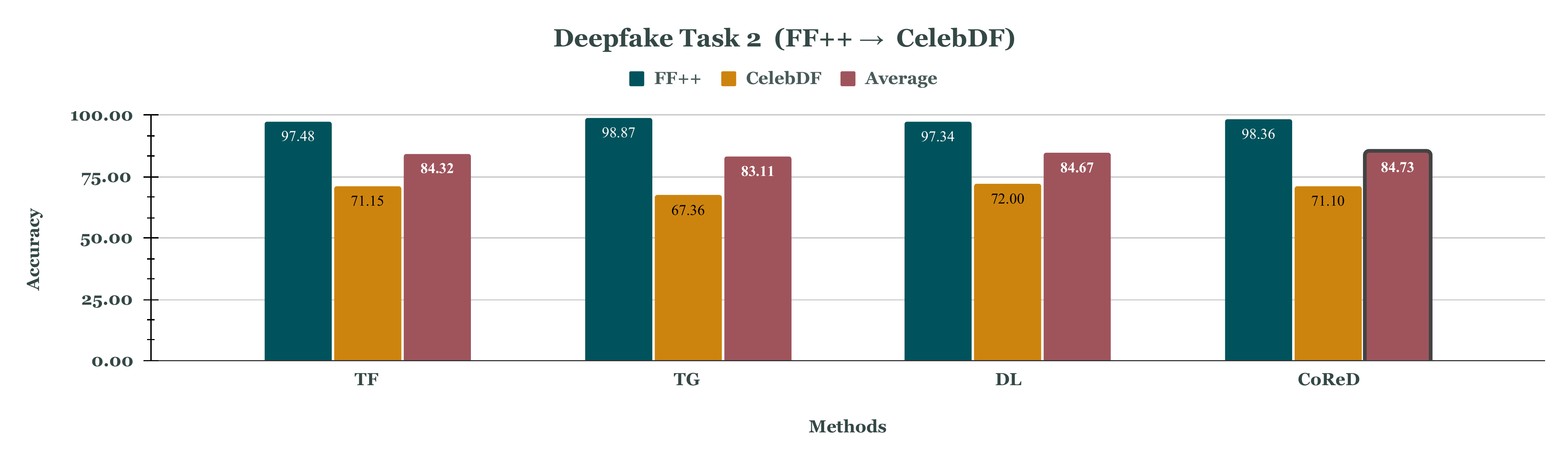}
    \caption{Task 2 results for deepfake video detection using FF++ and CelebDF datasets. In this case, we use FF++ for Task 1 and CelebDF dataset for Task 2. All the methods perform relatively well. These results are similar to the results for FF++ only Task 2 (see Fig.~\ref{fig:AUC_TASK2}). On Average, CoReD outperforms other baselines but with a very small margin (outlined in black; the rightmost bar chart).}
    \label{fig:AUC_TASK4}
    % \vspace{25pt}
\end{figure*}

\begin{table*}[!hbt]
\centering
\caption{We compare the adaptability of CoReD with the DL method by applying continual learning to two mixed datasets. After continual learning to two mixed target datasets (HQ), we can observe that CoReD consistently performs better than DL for mixed target datasets. The best performer is highlighted with boldface.}
\label{tab:HQ_FF_mix_twodata}
\resizebox{\linewidth}{!}{%
\arrayrulecolor{black}
\begin{tabular}{l|cc|cc|cc|cc|cc|cc} 
\toprule
\textbf{Dataset} & \multicolumn{2}{c|}{\textbf{DF $\rightarrow$ {}F2F\&FS}} & \multicolumn{2}{c|}{\textbf{DF $\rightarrow$ (NT \& FS)}} & \multicolumn{2}{c|}{\textbf{NT $\rightarrow$ (DF \& F2F)}} & \multicolumn{2}{c|}{\textbf{FS $\rightarrow$ (DF \& F2F)}} & \multicolumn{2}{c|}{\textbf{F2F $\rightarrow$ (DF \& FS)}} & \multicolumn{2}{c}{\textbf{F2F $\rightarrow$ (NT \& FS)}} \\ 
\hline
\textbf{Domain} & DL & {\textbf{CoReD}} & DL & {\textbf{CoReD}} & DL & {\textbf{CoReD}} & DL & {\textbf{CoReD}} & DL & {\textbf{CoReD}} & DL & {\textbf{CoReD}} \\ 
\hline
Source & 82.25 & 95.03 & 90.27 & 97.40 & 84.75 & 92.69 & 80.85 & 91.97 & 95.08 & 93.89 & 91.60 & 94.78 \\ 
\arrayrulecolor{black}\hline
Target1 & 73.23 & 79.68 & 84.20 & 87.97 & 78.55 & 84.82 & 84.28 & 85.16 & 89.81 & 91.79 & 89.20 & 92.93 \\ 
\hline
Target2 & 63.21 & 70.74 & 57.74 & 70.68 & 75.40 & 79.36 & 73.42 & 80.93 & 85.83 & 89.01 & 78.57 & 82.28 \\ 
\arrayrulecolor{black}\hline
\textbf{Average} & 72.90 & \textbf{81.82} & 77.40 & \textbf{85.35} & 79.57 & \textbf{85.62} & 79.52 & \textbf{86.02} & 90.24 & \textbf{91.56} & 86.46 & \textbf{90.00} \\
\bottomrule
\end{tabular}
 }
%  \vspace{25pt}
\end{table*}

\begin{table*}[!hbt]
\centering
\caption{The results of ablation study comparing single memory storage block ($b=1$) with regular CoReD ($b=5$) for low-quality deepfake videos. The best performer is highlighted with boldface.}
\label{tab:ablation_oneavg_HQ}
\resizebox{\linewidth}{!}{%
\begin{tabular}{l|cc|cc|cc|cc|cc|cc} 
\toprule
 \textbf{Method} & \multicolumn{2}{c|}{\textbf{DF $\rightarrow$ F2F}} & \multicolumn{2}{c|}{\textbf{DF $\rightarrow$ NT}} & \multicolumn{2}{c|}{\textbf{F2F $\rightarrow$ FS}} & \multicolumn{2}{c|}{\textbf{F2F $\rightarrow$ NT}} & \multicolumn{2}{c|}{\textbf{FS $\rightarrow$ DF}} & \multicolumn{2}{c}{\textbf{FS $\rightarrow$ F2F}} \\ 
\hline
\textbf{Domain} & \multicolumn{1}{c}{\textbf{CoReD}} & \multicolumn{1}{c|}{\begin{tabular}[c]{@{}c@{}}CoReD\\(One Block)\end{tabular}} & \multicolumn{1}{c}{\textbf{\textbf{CoReD}}} & \multicolumn{1}{c|}{\begin{tabular}[c]{@{}c@{}}CoReD\\(One Block)\end{tabular}} & \multicolumn{1}{c}{\textbf{\textbf{CoReD}}} & \multicolumn{1}{c|}{\begin{tabular}[c]{@{}c@{}}CoReD\\(One Block)\end{tabular}} & \multicolumn{1}{c}{\textbf{\textbf{CoReD}}} & \multicolumn{1}{c|}{\begin{tabular}[c]{@{}c@{}}CoReD\\(One Block)\end{tabular}} & \multicolumn{1}{c}{\textbf{\textbf{CoReD}}} & \multicolumn{1}{c|}{\begin{tabular}[c]{@{}c@{}}CoReD\\(One Block)\end{tabular}} & \multicolumn{1}{c}{\textbf{\textbf{CoReD}}} & \multicolumn{1}{c}{\begin{tabular}[c]{@{}c@{}}CoReD\\(One Block)\end{tabular}} \\ 
\hline
Source & 91.20 & 82.14 & 90.56 & 85.45 & 82.03 & 79.56 & 82.85 & 80.84 & 85.93 & 82.96 & 81.78 & 75.23 \\ 
\hline
Target & 62.09 & 66.64 & 83.38 & 82.35 & 68.79 & 58.92 & 83.87 & 83.69 & 65.78 & 63.96 & 64.45 & 61.86 \\ 
\hline
\textbf{Average} & \textbf{76.65} & 74.39 & \textbf{86.97} & 83.90 & \textbf{75.41} & 69.24 & \textbf{83.36} & 82.27 & \textbf{75.86} & 73.46 & \textbf{73.12} & 68.55 \\
\bottomrule
\end{tabular}
}
% \vspace{25pt}
\end{table*}

\begin{table*}[!hbt]
\centering
\caption{The results of ablation study comparing single storage ($b=1$) with regular CoReD ($b=5$) for high-quality deepfake videos. The best performer is highlighted with boldface.}
\label{tab:ablation_oneavg_LQ}
\resizebox{\linewidth}{!}{%
\begin{tabular}{l|cc|cc|cc|cc|cc|cc} 
\toprule
\textbf{Method} & \multicolumn{2}{c|}{\textbf{DF $\rightarrow$ F2F}} & \multicolumn{2}{c|}{\textbf{DF $\rightarrow$ FS}} & \multicolumn{2}{c|}{\textbf{F2F $\rightarrow$ NT}} & \multicolumn{2}{c|}{\textbf{F2F $\rightarrow$ FS}} & \multicolumn{2}{c|}{\textbf{FS $\rightarrow$ DF}} & \multicolumn{2}{c}{\textbf{FS $\rightarrow$ F2F}} \\ 
\hline
\textbf{Domain} & \multicolumn{1}{c}{\textbf{CoReD}} & \multicolumn{1}{c|}{\begin{tabular}[c]{@{}c@{}}CoReD\\(One Block)\end{tabular}} & \multicolumn{1}{c}{\textbf{CoReD}} & \multicolumn{1}{c|}{\begin{tabular}[c]{@{}c@{}}CoReD\\(One Block)\end{tabular}} & \multicolumn{1}{c}{\textbf{CoReD}} & \multicolumn{1}{c|}{\begin{tabular}[c]{@{}c@{}}CoReD\\(One Block)\end{tabular}} & \multicolumn{1}{c}{\textbf{CoReD}} & \multicolumn{1}{c|}{\begin{tabular}[c]{@{}c@{}}CoReD\\(One Block)\end{tabular}} & \multicolumn{1}{c}{\textbf{CoReD}} & \multicolumn{1}{c|}{\begin{tabular}[c]{@{}c@{}}CoReD\\(One Block)\end{tabular}} & \multicolumn{1}{c}{\textbf{CoReD}} & \multicolumn{1}{c}{\begin{tabular}[c]{@{}c@{}}CoReD\\(One Block)\end{tabular}} \\ 
\hline
Source & 95.58 & 84.56 & 88.60 & 88.24 & 98.09 & 96.82 & 93.36 & 81.79 & 92.30 & 61.71 & 96.01 & 89.89 \\ 
\hline
Target & 84.24 & 81.01 & 76.23 & 74.92 & 88.90 & 91.76 & 80.63 & 78.31 & 86.45 & 85.10 & 88.64 & 85.95 \\ 
\hline
\textbf{Average} & \textbf{89.91} & 82.79 & \textbf{82.42} & 81.58 & 93.50 & \textbf{94.29} & \textbf{87.00} & 80.05 & \textbf{89.38} & 73.41 & \textbf{92.33} & 87.92 \\
\bottomrule
\end{tabular}
}
% \vspace{25pt}
\end{table*}

\begin{figure*}[!hbt]
    \centering
    \includegraphics[clip, trim=20pt 18pt 85pt 22pt, width=1\linewidth]{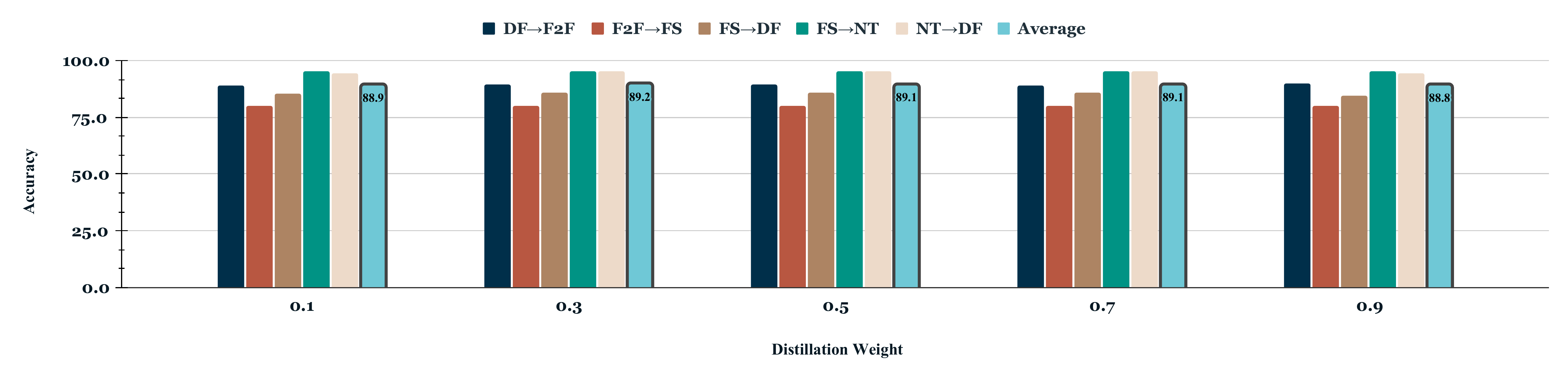}
    \caption{Ablation study different values of distillation weight for CoReD. We evaluated five different alpha values for Knowledge Distillation inside CoReD. As we can see, distillation weight's value does not significantly impact the overall model performance, demonstrating that our CoReD model is significantly stable. Furthermore, distillation weight = 0.3 gives the best performance on average. Therefore, we used it for the rest of our experiments (i.e., Task 3, $\dots$, Task N).}
    \label{fig:AUC_TASK4}
\end{figure*}

\end{document}